\documentclass[letterpaper, 10 pt, conference]{ieeeconf}  

\IEEEoverridecommandlockouts                              

\let\labelindent\relax
\usepackage[inline]{enumitem} 
\usepackage{graphicx}
\usepackage{subcaption}
\usepackage{rotating}
\usepackage{booktabs}
\usepackage{epsfig} 
\usepackage{amsmath} 
\usepackage{amssymb}  

\usepackage{url}
\usepackage{color}

\usepackage{multirow}
\usepackage{siunitx}
\usepackage{makecell}

\makeatletter
\let\NAT@parse\undefined
\makeatother
\def\secref#1{Sec.\ref{#1}}
\def\figref#1{Fig.\ref{#1}}
\def\tabref#1{Tab.\ref{#1}}
\def\eqref#1{Eq.(\ref{#1})}

\renewcommand{\vec}[1]{\mathbf{#1}}

\DeclareMathAlphabet\mathbfcal{OMS}{cmsy}{b}{n}
\newcommand{\gt}{ground-truth }
\usepackage[numbers]{natbib}
\newcolumntype{L}[1]{>{\raggedright\arraybackslash}p{#1}}
\newcolumntype{C}[1]{>{\centering\arraybackslash}p{#1}}
\newcolumntype{R}[1]{>{\raggedleft\arraybackslash}p{#1}}
\title{\LARGE \bf
DeepTemporalSeg: Temporally Consistent Semantic Segmentation of 3D LiDAR Scans
}

\author{Ayush Dewan$^1$ \and Wolfram Burgard$^{1,2}$  
\thanks{$^1$Department of Computer Science, University of Freiburg, Germany.}\thanks{$^2$Toyota Research Institute, Los Altos, USA.}}

\begin{document}

\maketitle
\thispagestyle{empty}
\pagestyle{empty}

\begin{abstract}
Understanding the semantic characteristics of the environment is a key enabler for autonomous robot operation. In this paper, we propose a deep convolutional neural network (DCNN) for semantic segmentation of a LiDAR scan into the classes \textit{car}, \textit{pedestrian} and \textit{bicyclist}. This architecture is based on dense blocks and efficiently utilizes depth separable convolutions to limit the number of parameters while still maintaining the state-of-the-art performance. To make the predictions from the DCNN temporally consistent, we propose a Bayes filter based method. This method uses the predictions from the neural network to recursively estimate the current semantic state of a point in a scan. This recursive estimation uses the knowledge gained from previous scans, thereby making the predictions temporally consistent and robust towards isolated erroneous predictions. We compare the performance of our proposed architecture with other state-of-the-art neural network architectures and report substantial improvement. For the proposed Bayes filter approach, we shows results on various sequences in the KITTI tracking benchmark.
\end{abstract}

\section{Introduction}

In the last decade, the research towards self-driving cars has picked up a staggering pace. The main objective of this technology is to make our roads safer than ever before~\cite{waymo-mission}. A key ingredient to realize the goals of autonomous vehicles is a robust perception system, where the main objective is to understand the environment in which the robot is operating, through a variety of sensors that a robot is endowed with. In this paper, we focus of semantic scene understanding of urban outdoor environment using 3D LiDAR scans. Semantic understanding is crucial, as it paves the way for robust visual localization~\cite{naseer2017semantics,radwan2018vlocnet++}, efficient mapping~\cite{ruchti18icra}, among several other tasks.

In this paper, we propose a deep convolutional neural network (DCNN) architecture for the task of semantic segmentation of a 3D LiDAR scan into the following semantic categories: \textit{car}, \textit{pedestrian} and \textit{bicyclist}. Our proposed architecture is based on dense blocks~\cite{huang2016densely}. To reduce the number of parameters, we replace the standard convolution layers with depth separable convolution layers~\cite{chollet2017xception} for dense blocks in the decoder. This allows us to reduce the number of parameters by a significant amount while still having competitive performance.

\begin{figure}[]

     \centering
 	  \begin{subfigure}[]{0.168\textwidth}
      \includegraphics[width=0.65\linewidth]{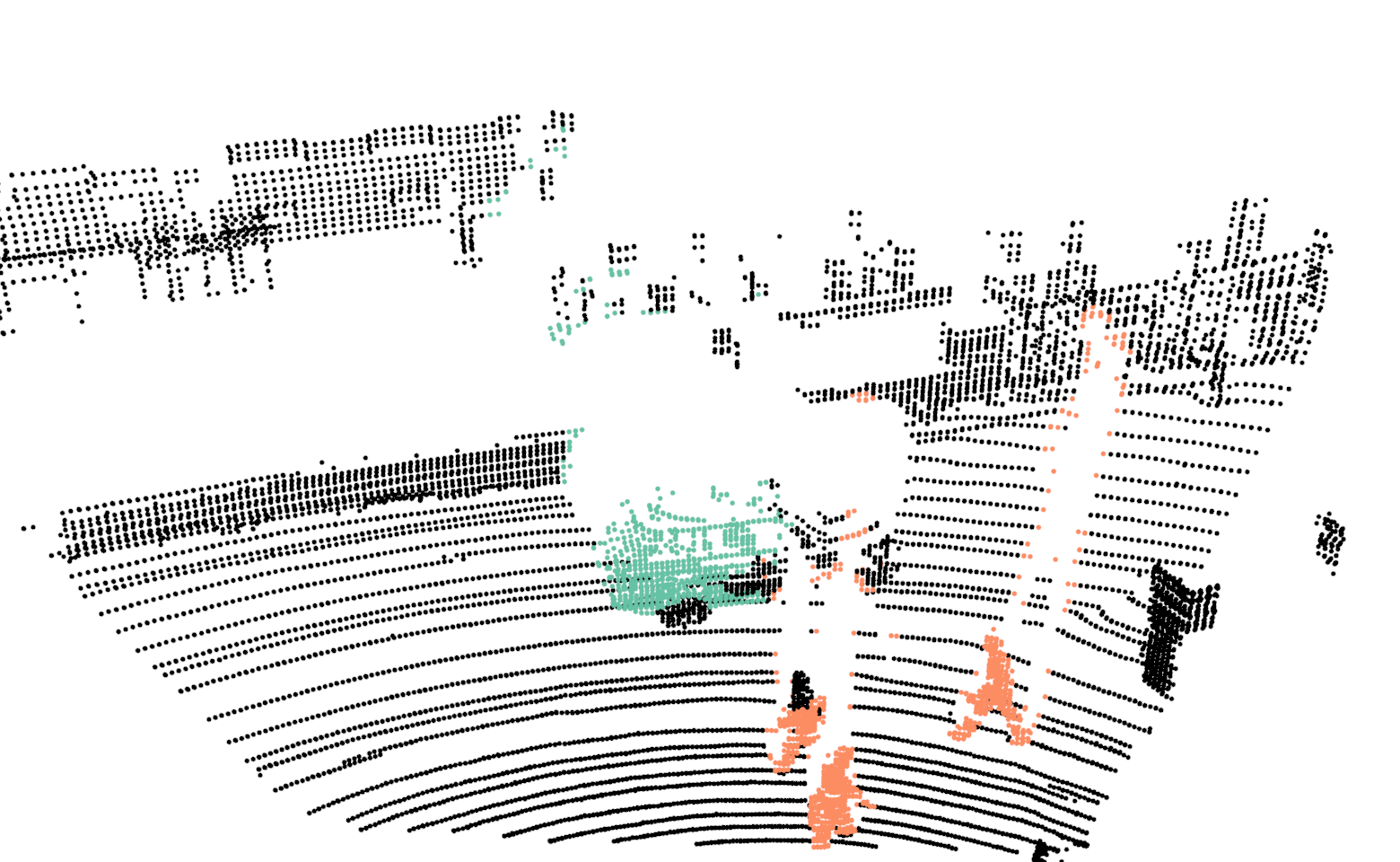}
    \caption{}
    \label{fig:resnet_corr}
  \end{subfigure}%
  \centering
  \begin{subfigure}[]{0.168\textwidth}
    \centering
    \includegraphics[width=0.65\linewidth]{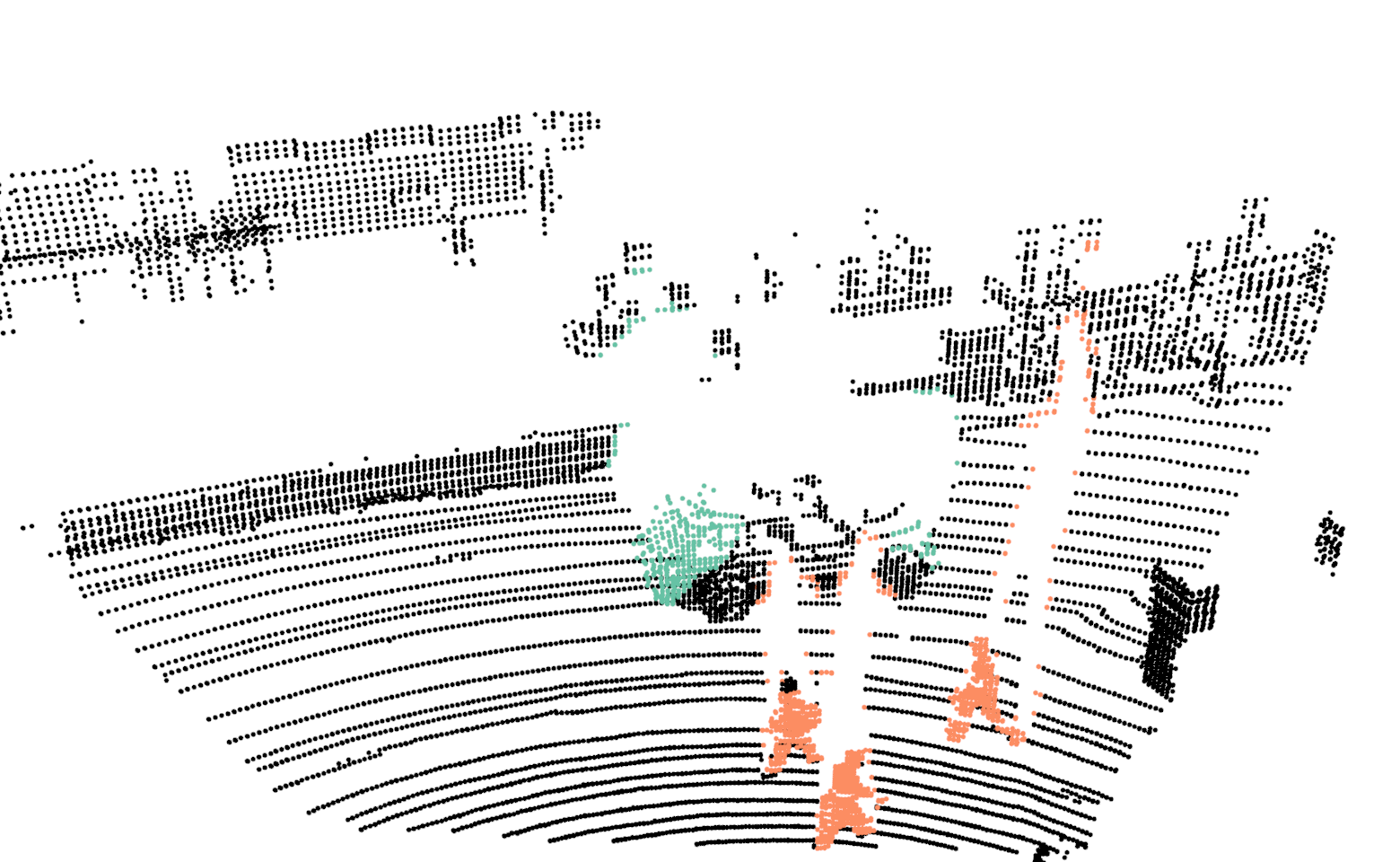}
    \caption{}
    \label{fig:resnet_tsne}
    \end{subfigure}%
      \centering
 	  \begin{subfigure}[]{0.168\textwidth}
      \includegraphics[width=0.65\linewidth]{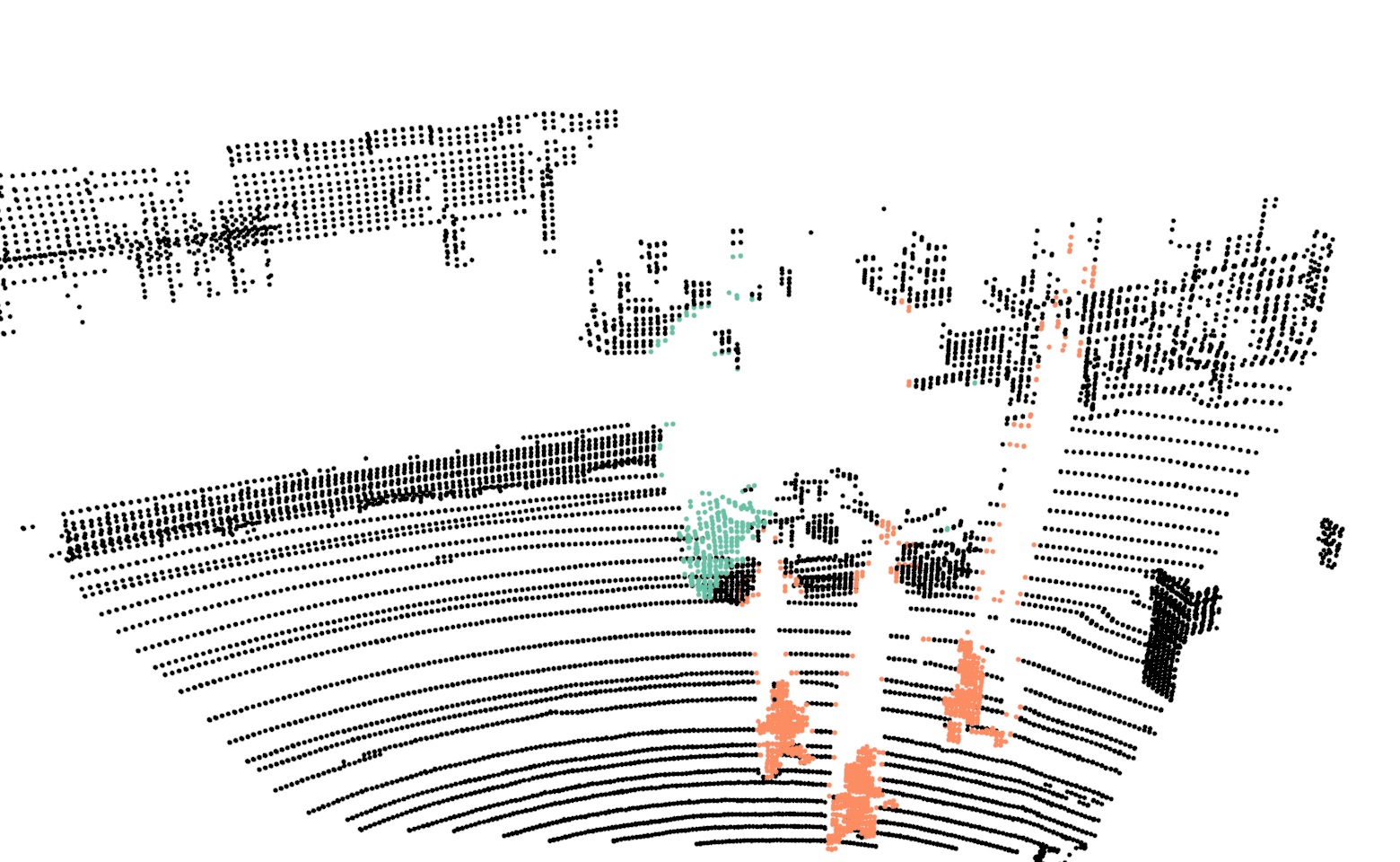}
    \caption{}
    \label{fig:matchnet_corr}
  \end{subfigure}%
  \\
       \centering
 	  \begin{subfigure}[]{0.168\textwidth}
      \includegraphics[width=0.65\linewidth]{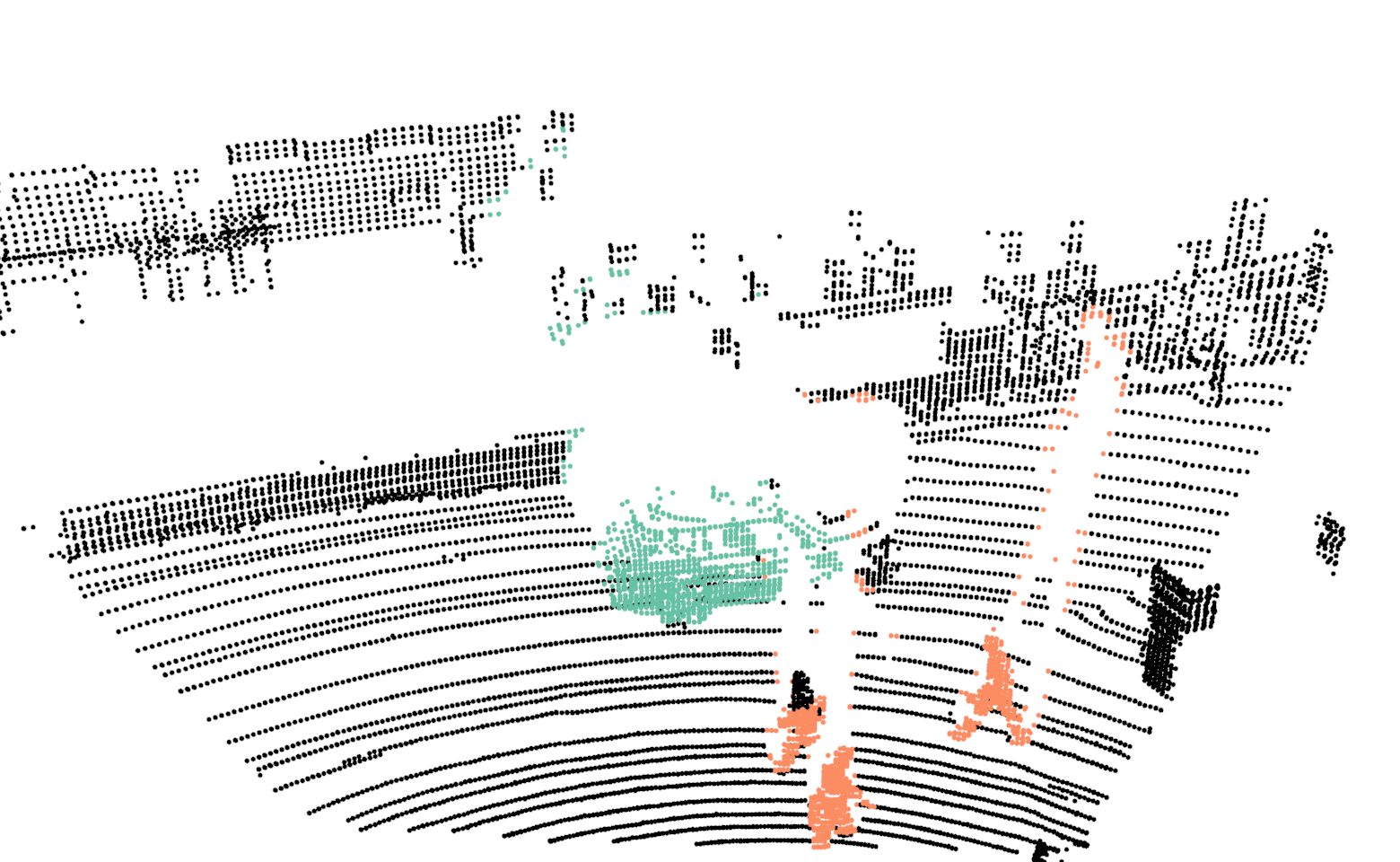}
    \caption{}
    \label{fig:resnet_corr}
  \end{subfigure}%
  \centering
  \begin{subfigure}[]{0.168\textwidth}
    \centering
    \includegraphics[width=0.65\linewidth]{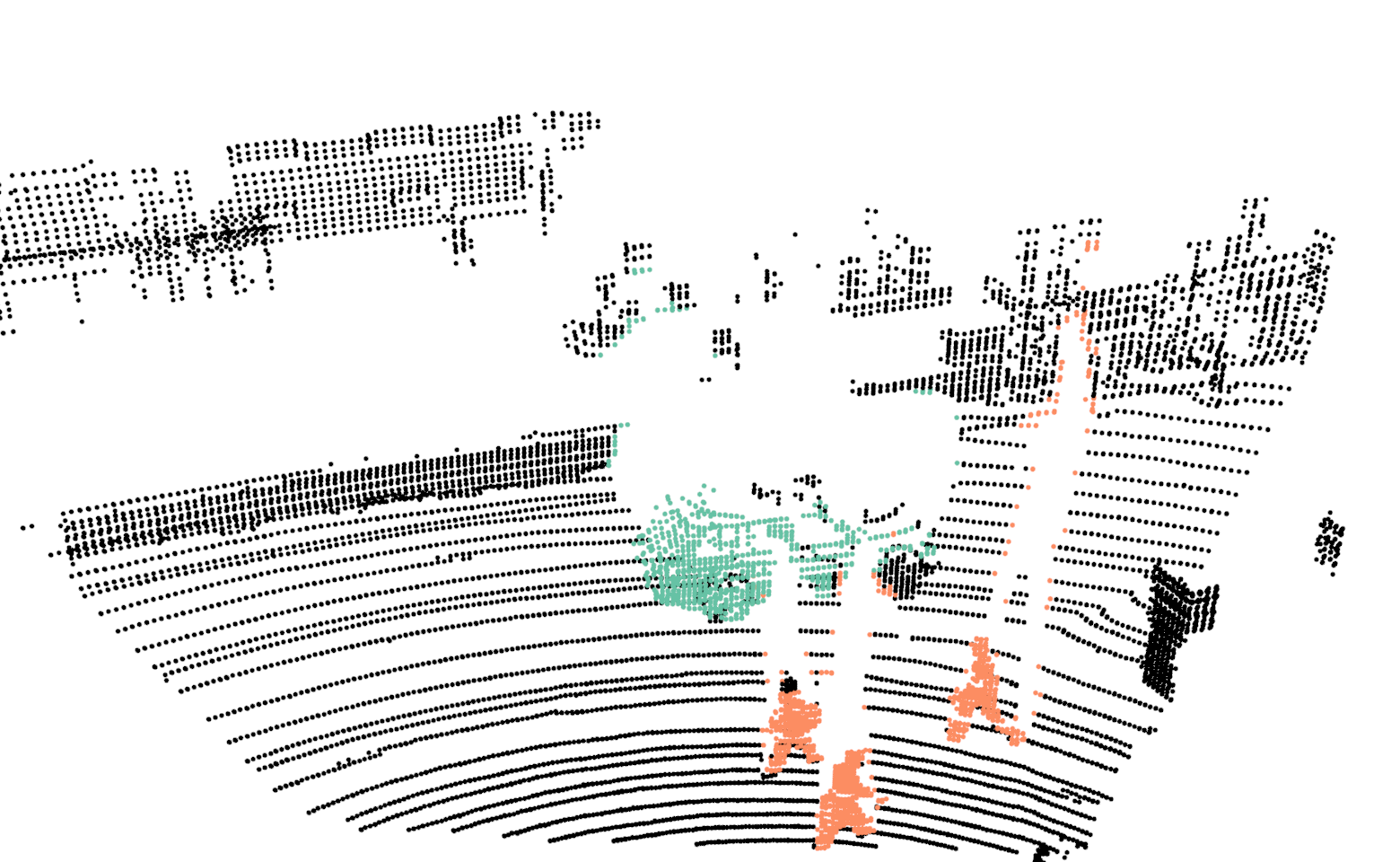}
    \caption{}
    \label{fig:resnet_tsne}
    \end{subfigure}%
      \centering
 	  \begin{subfigure}[]{0.168\textwidth}
      \includegraphics[width=0.65\linewidth]{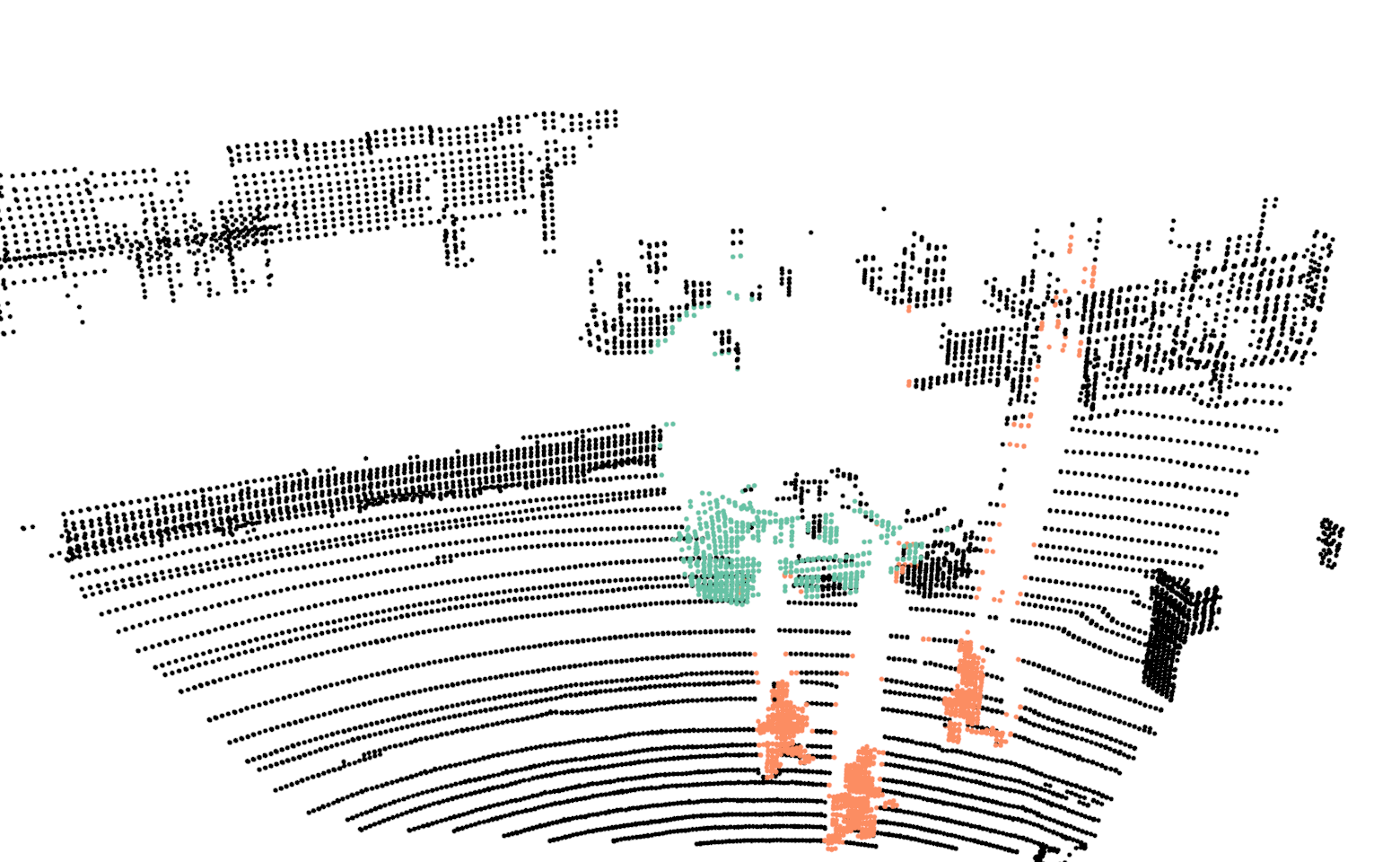}
    \caption{}
    \label{fig:matchnet_corr}
  \end{subfigure}%
   
      \caption{Illustration of semantic segmentation with our proposed methods. In the top row ((a)-(c)), we show the output of our proposed DCNN, for three consecutive scans. In the top left image, points on a car (color green) are correctly classified, but in subsequent scans, points on the same car are partially ((b)-(c)) misclassified as background. In the bottom row, we show the output of our proposed binary Bayes filter. For all the scans, points on the same car are correctly classified.}
\label{fig:object_bayes_filter}    
\end{figure}   
Standard DCNN architectures treat each example independently and do not use any previous or prior information. Especially in the case of perception in robotics, the data is sequential. To leverage over this sequential nature of information, we propose a Bayes filter approach for making our segmentation results temporally consistent. More concretely, we use a Bayes filter with a \textit{static} state, where static in this context means that transition between different states is unlikely, which is true for semantic classes. This approach neatly combines the current prediction of the neural network, with the information accumulated from previous scans. In our previous work~\cite{dewan2017deep}, such an approach was part of our method to classify points in a 3D LiDAR scan as non-movable, movable and dynamic. We illustrated its advantages through qualitative results. In this paper, we thoroughly analyze our method by evaluating our approach on various sequences of KITTI tracking benchmark and report both qualitative and quantitative results.

The main contributions of this paper include a DCNN for semantic segmentation of LiDAR scans into the classes: \textit{car}, \textit{pedestrian} and \textit{bicyclist}. We compare our DCNN with state-of-the-art DCNNs~\cite{wu2018squeezeseg, wu2018squeezesegv2, wang2018pointseg}, proposed for the same task. To justify different architecture design choices and gain further insight towards them, we also present an ablation study. Our next contribution is a Bayes filter approach for making the predictions of the neural network temporally consistent. This approach leverages over the sequential nature of the input data stream and makes our segmentation system robust towards sporadic erroneous prediction. For comparison, we use our proposed architecture as a baseline method. The code for the proposed architecture, along the trained model and the dataset is available here \url{http://deep-temporal-seg.informatik.uni-freiburg.de/}.

\section{Related Work}
With the advent of deep neural networks, a significant progress has been made towards solving a variety of tasks, including the task of semantic segmentation. Regarding 2D images, a plethora of research has been done in last few years~\cite{long_shelhamer_fcn, ronneberger2015u,kendall2015,jegou2017one,chen2017rethinking}, pushing the boundary of state-of-the-art results to the limit. A similar progress has not been in the field of semantic segmentation of 3D pointcloud data due to inherent differences in the two data modalities. In the case of 2D images, the input data to the network is fixed but in the case of 3D data, multiple representations are possible. Regarding the current task, the most commonly used representation are either a collection of 3D points or projecting the pointcloud on a 2D image. For the first representation, the PointNet and PointNet++ architecture proposed by~\citeauthor{qi2017pointnet}~\cite{qi2017pointnet, qi2017pointnet++} is a popular choice for learning from unordered pointcloud. They have shown results primarily on indoor sequence for the data collected from RGB-D sensors. In our case, we use a LiDAR scanner for segmentation of urban outdoor environments. The data from LiDAR scanner is sparser in comparison to the RGB-D sensor and the outdoor environment is more spread out in comparison to confined indoor spaces. In our case, we use the second representation i.e. projecting the 3D LiDAR scan on to a 2D image. This allows us to represent a LiDAR scan in a compact fashion and furthermore the advancements made in the field of semantic segmentation using 2D images can be used as well.

Focusing on the task of semantic segmentation using 2D images, one of the initial architectures was proposed by~\citeauthor{long_shelhamer_fcn}~\cite{long_shelhamer_fcn}. They proposed an encoder-decoder style, fully convolutional network (FCN) architecture and other architectures since then have followed the same paradigm.~\citeauthor{jegou2017one}~\cite{jegou2017one} proposed a dense block based DCNN for the task of semantic segmentation. The main differences between our DCNN and theirs is that we use depth separable convolution layers for dense blocks in the decoder. To down-sample the feature maps they proposed a transition down block comprising of a composite function implementing different operations. We replace this block with a single max-pooling operation and show that instead of a composite function, this single operation is sufficient. In the presented ablation study, we justify these proposed changes. 

We compare our proposed architecture with the architectures proposed in~\cite{wu2018squeezeseg, wu2018squeezesegv2, wang2018pointseg}. The first architecture proposed by ~\citeauthor{wu2018squeezeseg}~\cite{wu2018squeezeseg} is based on the SqueezeNet~\cite{iandola2016squeezenet} architecture. They use \textit{fire} modules, which first involves squeezing the feature maps using $1\times 1$ filters and then expanding these squeezed feature maps in parallel using filters of size $1\times 1$ and $3\times 3$ and concatenating their outputs at the end. Using three max-pool layers they down-sample the feature maps only along the width dimension and to up-sample the feature maps they again use \textit{fire} modules in the decoder. Last layer of their neural network architecture is a recurrent CRF and the complete architecture is trained end-to-end. In our ablation study, we compare with their proposed down-sampling technique.

They further improve this architecture in~\cite{wu2018squeezesegv2} by using a binary mask as an additional input channel. This mask indicates existence of a LiDAR measurement corresponding to a pixel location. Along this they also introduce a novel context aggregation module to limit the error introduced by missing LiDAR measurements and furthermore in order to tackle the class imbalancing problem they use focal loss~\cite{lin2017focal} for training their DCNN. The last method we compare with is the DCNN proposed by~\citeauthor{wang2018pointseg}~\cite{wang2018pointseg}. Similar to the neural network architectures proposed by~\citeauthor{wu2018squeezeseg}~\cite{wu2018squeezeseg}, their network architecture is also based on SqueezeNet. They also use Squeeze Excitation blocks~\cite{hu2018squeeze} after the initial \textit{fire} modules and at the end of the encoder use an \textit{enlargement} layer which is based on the Atrous Spatial Pyramid Pooling~\cite{chen2018deeplab}.

\section{Network Architecture}
\label{sec:seg_nn}
\begin{figure*}[]
  \centering
    \includegraphics[width=0.70\textwidth]{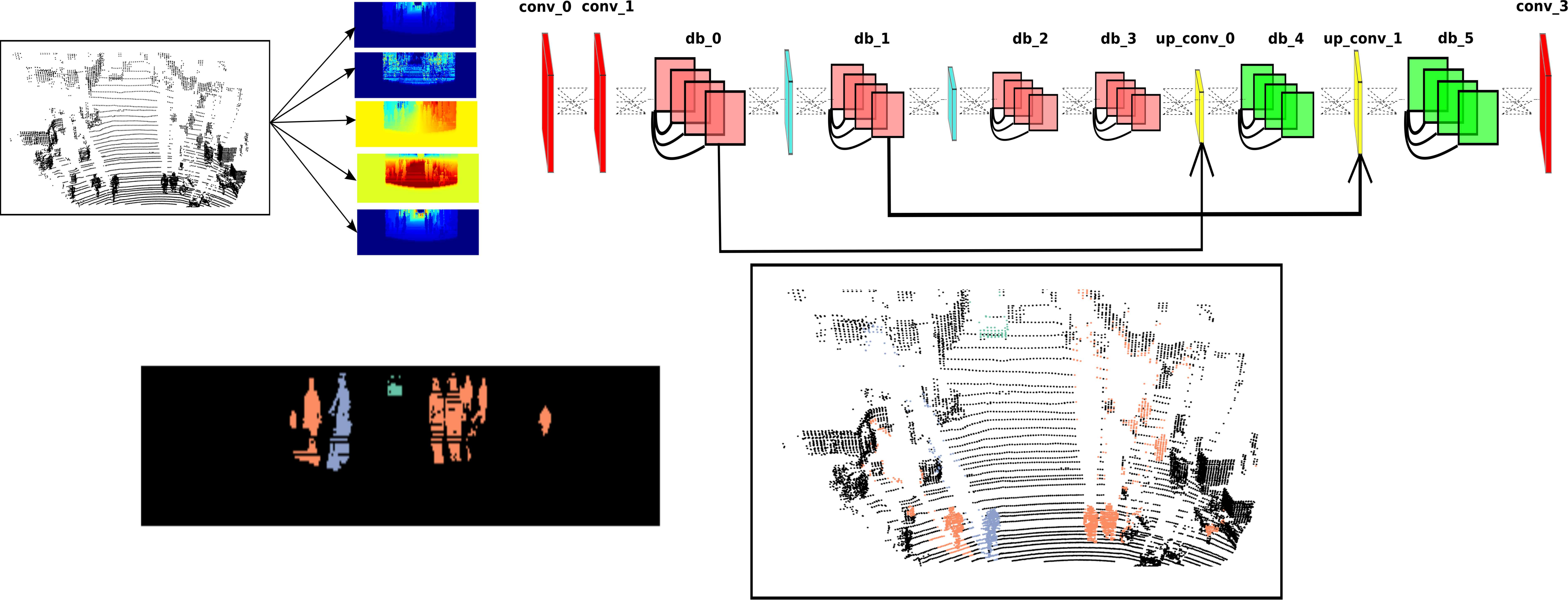}
    \caption[Proposed semantic segmentation framework]{Our proposed semantic segmentation framework. In the first step we project a LiDAR scan onto five 2D images and encode the following modalities: depth, surface reflectance intensity, 3D coordinates(x, y, z). These images are then stacked together, fed into the proposed CNN architecture, and the output is the predicted segmentation mask (bottom left). The segmentation information is then projected back to the scan to infer the semantic labels for each point in the scan. In the encoder we have two convolution layers (conv\_0 and conv\_1), two max-pooling layers and four dense blocks (db\_0, db\_1, db\_2 and db\_3). In the decoder we use two up-convolution layers, two dense blocks (db\_4, db\_5) with depth separable convolution and one convolution layer (conv\_2). 
}
\label{fig:seg_summary}
\end{figure*}
In~\figref{fig:seg_summary} we illustrate the complete framework for semantic segmentation of a LiDAR scan. The first step is to project the scan onto different 2D images and each such image encodes a specific modality. These images are then stacked together and are passed through our proposed DCNN for semantic segmentation. The segmentation mask predicted by the DCNN is then projected back to the LiDAR scan to infer pointwise semantic labels.

For the task of semantic segmentation we a propose a novel fully convolutional DCNN architecture called DBLiDARNet. Our architecture is based on dense blocks and is shown in~\figref{fig:seg_summary}. Similar to other DCNN architecture proposed for the task of semantic segmentation~\cite{jegou2017one, long_shelhamer_fcn, ronneberger2015u}, our network is also comprised of an encoder for learning the features required for the task while down-sampling the feature map size and a decoder to up-sample the feature maps so that the last hidden layer has the same spatial resolution as the input image. In the encoder, we have two convolution layers (conv\_0 and conv\_1), four dense blocks (db\_0, db\_1, db\_2 and db\_3) and two max-pooling layers to down-sample the feature maps $4\times$ in comparison to the spatial resolution of the input image. In the decoder we use two up-convolution layers to up-sample the feature maps and use two more dense blocks with depth separable convolution layers. To limit the number of learnable parameters in the decoder, similar to the architecture proposed by~\citeauthor{jegou2017one}~\cite{jegou2017one}, in our proposed architecture the input to the up-convolution layers is the feature maps learned by the dense block prior to the up-convolution layer instead of all the features maps learned till that point. For instance the input to the layer up\_conv\_0 is only the feature maps learned by the dense block db\_3. To recapture the information lost during up-sampling we use skip connections to concatenate the feature maps from the encoder to the output of the up-convolution layers. The complete details regarding the dimensions of each layer or block and different associated hyper-parameters is reported in~\tabref{tab:layer_details}. \begin{table}[]
 \centering
 \caption{Architecture}
 \begin{tabular}{|c|l|c|c|}
 \hline
 Layer name&Dimension (H$\times$W$\times$C)& Repetition& Depth separable\\\hline
 conv\_0&$64\times512\times48$&-&No\\
 conv\_1&$64\times512\times48$&-&No\\
 db\_0&$64\times512\times144$&6&No\\
 db\_1&$32\times256\times272$&8&No\\
 db\_2&$16\times128\times432$&10&No\\
 db\_3&$16\times128\times240$&15&Yes\\
 up\_conv\_0&$32\times256\times240$&-&No\\
 db\_4&$32\times256\times128$&8&Yes\\
 up\_conv\_1&$64\times512\times128$&-&No\\
 db\_5&$64\times512\times96$&6&Yes\\
 conv\_2&$64\times512\times4$&-&No\\
 \hline
 \end{tabular}
  \label{tab:layer_details}
\end{table}
\subsection{Training}
\label{sec:seg_training}
Our complete network architecture is implemented in
TensorFlow~\cite{abadi2016tensorflow}. We use the dataset provided by~\citeauthor{wu2018squeezeseg}~\cite{wu2018squeezeseg}. We use softmax cross-entropy loss and use the Adam optimizer~\cite{kingma2014adam} with a learning rate of $1e^{-4}$, weight decay of $5e^{-4}$ and batch size of $2$. Among the three classes, the point measurements from cars is significantly more than the measurements from either pedestrians or bicyclists, mainly because of the inherent difference in the size of the geometrical structure. This leads to the problem of class imbalancing, where some classes in the training data overwhelm the classes which are under represented. To tackle this we use a weight balancing technique and assign larger weights to points belonging to the class \textit{pedestrian} and \textit{bicyclist} in comparison to points belonging to the class \textit{car} and \textit{background}.

\section{Bayes Filter Method}
In the~\secref{sec:seg_nn}, we proposed a novel DCNN architecture for semantic segmentation of a LiDAR scan into different categories. The output of the network is the predicted softmax probabilities of a point in a scan belonging to different categories. Since this prediction is performed independently for different scans, in this section we introduce a novel Bayes Filter approach to make our pointwise prediction temporally consistent. This approach assumes the scans are sequential with significant overlap and the objective is to leverage over this sequential nature of information and make our prediction robust to isolated erroneous predictions from the neural network. 

The semantic state of a point is \textit{static}, i.e. it remains same over time and transition between these states is unlikely. For each point, we use three separate binary Bayes filters with \textit{static} state, to estimate the belief for each class independently. To estimate the belief for a class $c$, for a point $\vec{p}^t \in \mathbb{R}^3$ in a scan at time $t$ , we first define a binary random state variale $O^t_c = \lbrace 0,1\rbrace$, where $O^t_c = 1$ indicates that the point belongs to the class $c$ and $O^t_c = 0$ indicates the opposite. Without loss of generality, from now on, we would write $Bel(O^t_c = 1)$ as $Bel(O^t_c)$ and $Bel(O^t_c = 0)$ as $Bel(\neg O^t_c)$. The current belief $Bel(O^t_c)$ depends only on the predictions of the neural network, $\xi^{1:t}_c$, for the class $c$ as shown in~\eqref{eq:belief_semantic}.
\begin{equation}
Bel(O^t_c) = P(O^t_c \mid \xi^{1:t}_c),\label{eq:belief_semantic}
\end{equation} 
where, $\xi^{1:t}_c$ are softmax scores for the class $c$. We define such binary random variables for each class and estimate the belief for each class independently. Using Bayes rule and Markov assumption we can rewrite the~\eqref{eq:belief_semantic} as
following,
\begin{equation}
P(O^t_c \mid \xi^{1:t}_c) = \frac{P(\xi^t_c\mid O^t_c)P(O^t_c\mid \xi^{1:t-1})}{P(\xi^t \mid \xi^{1:t-1})}.\label{eq:belief_semantic_bayes_markov}
\end{equation} 
Using Bayes rule for the term $P(\xi^t_c\mid O^t_c)$,~\eqref{eq:belief_semantic_bayes_markov} can be modified as following,
\begin{equation}
P(O^t_c \mid \xi^{1:t}_c) = \frac{P(O^t_c\mid \xi^t_c)P(\xi^t_c)P(O^t_c\mid \xi^{1:t-1})}{P(O^t_c)P(\xi^t \mid \xi^{1:t-1})}.\label{eq:belief_semantic_bayes_markov_bayes}
\end{equation} 
Similarly, $P(\neg O^t_c \mid \xi^{1:t}_c)$ can be written as,
\begin{equation}
P(\neg O^t_c \mid \xi^{1:t}_c) = \frac{P(\neg O^t_c\mid \xi^t_c)P(\xi^t_c)P(\neg O^t_c\mid \xi^{1:t-1})}{P(\neg O^t_c)P(\xi^t \mid \xi^{1:t-1})}.\label{eq:belief_neg_semantic_bayes_markov_bayes}
\end{equation}

We now introduce the log odds notation, where odds of an event $x$ is defined in~\eqref{eq:odds} and the log odds are defined in~\eqref{eq:log_odds}
\begin{align}
\frac{p(x)}{\neg p(x)}&=\frac{p(x)}{1 - p(x)},\label{eq:odds}\\
l(x)&=~\text{log}\frac{p(x)}{1 - p(x)}.\label{eq:log_odds}
\end{align}

The odds for a point $\vec{p}^t$ having the semantic class $c$ can be estimated by dividing~\eqref{eq:belief_semantic_bayes_markov_bayes} with~\eqref{eq:belief_neg_semantic_bayes_markov_bayes}. The odds is defined in~\eqref{eq:class_odds} and the log odds are defined in~\eqref{eq:class_log_odds}, 
\begin{align}
\frac{P(O^t_c \mid \xi^{1:t}_c)}{P(\neg O^t_c \mid \xi^{1:t}_c)}&=\frac{P(O^t_c\mid \xi^t_c)}{P(\neg O^t_c\mid \xi^t_c)}\frac{P(O^t_c\mid \xi^{1:t-1})}{\neg P(O^t_c\mid \xi^{1:t-1})}\frac{P(\neg O^t_c)}{P(O^t_c)},\label{eq:class_odds}\\
&=\frac{P(O^t_c\mid \xi^t_c)}{1-P(O^t_c\mid \xi^t_c)}\frac{P(O^t_c\mid \xi^{1:t-1})}{1-P(O^t_c\mid \xi^{1:t-1})}\\
&\phantom{{}=1} \frac{1-P(O^t_c)}{P(O^t_c)}, \notag\\
l_t(O_c^t)&=~\text{log}\frac{P(O^t_c\mid \xi^t_c)}{1-P(O^t_c\mid \xi^t_c)} + l_{t-1}(O_c) - l_0(O_c),\label{eq:class_log_odds}
\end{align}
where, the current measurement is defined as following,
\begin{equation}
P(O^t_c\mid \xi^t_c) = \xi^t_c\label{eq:nn_measurement}.
\end{equation}

In~\eqref{eq:class_log_odds}, $l_t(O_c^t)$ are the log odds for the belief at time $t$, the first term on the right side in~\eqref{eq:class_log_odds} are the log odds for the current measurement, $l_{t-1}(O_c)$ are the log odds for the previous belief and $l_0(O_c)$ are the log odds for the  initial belief. Through this formulation, our inference not only depends on the current measurement ($P(O^t_c\mid \xi^t_c)$), but also on the previous measurements, incorporated through the recursive term  $l_{t-1}(O_c)$. To enable this recursive behavior, data association between points in consecutive scans is required and for this we use our method of estimating pointwise motion proposed in~\cite{dewan2016rigid}. We perform data association by aligning scans using the estimated motion and choosing the nearest point on the basis of Euclidean distance as the corresponding point. As mentioned before, we estimate $l_t(O_c^t)$ for each class separately and for the inference we choose the class with the largest odds.

\section{Results}
\label{sec:segmentation_results}
\subsection{Network Architecture}
\begin{figure*}[]
  \centering
    \includegraphics[width=0.65\textwidth]{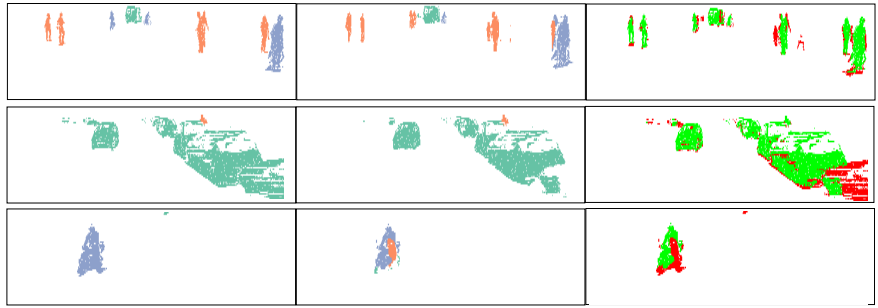}
    \caption[An illustration of the semantic segmentation results]{An illustration of the semantic segmentation results. In the left column we show the~\gt segmentation masks where points belonging to the class \textit{car}, \textit{pedestrian} and \textit{bicyclist} are show in color green, orange and blue respectively. In the middle column we show the predicted segmentation masks with the same color scheme as the~\gt masks. To clearly visualize the differences between the~\gt and predicted masks, in the last we show the correctly segmented points in green color and the misclassified points in color red. The top row illustrates the case where our proposed DCNN is able to successfully segment objects of different classes. The middle row shows a hard case, where a pedestrian is walking behind the cars and is heavily occluded and our method is still able to correctly segment the pedestrian. The bottom row illustrates a case where our method under performs. In some cases bicyclists are over segmented into the classes \textit{bicyclist} and \textit{pedestrian} due to  presence of a person in both classes.}
\label{fig:seg_results}
\end{figure*}

To evaluate our proposed DCNN, we use the test set from the dataset provided by~\citeauthor{wu2018squeezeseg}~\cite{wu2018squeezeseg}. We report class wise IoU and compare our results with two DCNN proposed by~\citeauthor{wu2018squeezeseg} (\cite{wu2018squeezeseg},\cite{wu2018squeezesegv2}) and the network architecture proposed by~\citeauthor{wang2018pointseg}~\cite{wang2018pointseg}. In~\figref{fig:seg_results}, we show qualitative semantic segmentation results. In~\tabref{tab:seg_IoU} we report the class wise IoU and mean IoU for different methods. Our proposed DCNN outperforms the existing state-of-the art DCNNs proposed for the same task and has a better IoU for all the three classes. In the case of \textit{pedestrian}, the increase in IoU is around 70\%, for the class \textit{bicyclist} the increase is around 17\%, with an overall increase in mean IoU by 16\%. These results indicate a remarkable improvement over the existing DCNNs proposed to solve the same task. 
Comparing the inter class performance, the highest IoU is achieved for the class \textit{car}, whereas the performance for \textit{pedestrian} and \textit{bicyclist} are comparable. Similar trend is evident for other methods as well. This difference in performance has three main reasons, firstly the number of instances of \textit{pedestrian} and \textit{bicyclist} is lesser in comparison to \textit{car}. Secondly, object in both these classes have a smaller size in comparison to cars and therefore the number of points sampled from their surface is significantly lower in comparison to points sampled from the surface of cars. Due to these reasons, these two classes are under represented and as mentioned before, we use weight balancing in order to have a large penalty for misclassifying points in these classes. The last reason is the over segmentation of points on a bicyclist into classes \textit{bicyclist} and \textit{pedestrian} as shown in~\figref{fig:seg_results}. This misclassification is not a common occurrence but still hampers the overall performance.


\begin{table}[]
 \centering
 \caption{A comparison with other DCNNs proposed for semantic segmentation of a LiDAR scan. For each method we report class wise and mean IoU}
\vspace{-0.1cm} 
 \begin{tabular}{|c|c|c|c|c|c|l|}
 \hline
 Method&Cars&Pedestrians&Bicyclists&meanIoU\\\hline
 SqueezeSeg~\cite{wu2018squeezeseg}&60.9&22.8&26.4&36.7\\
 SqueezeSeg w/ CRF~\cite{wu2018squeezeseg}&64.6&21.8&25.1&37.1\\
 PointSeg~\cite{wang2018pointseg}&67.4&19.2&32.7&39.7\\
 PointSeg w/ RANSAC~\cite{wang2018pointseg}&67.3&23.9&38.7&43.3\\
 SqueezeSegV2~\cite{wu2018squeezesegv2}&73.2&27.8&33.6&44.8\\
 DBLiDARNet (Ours)&\textbf{75.1}&\textbf{47.4}&\textbf{45.4}&\textbf{56.0}\\
  \hline
 \end{tabular}
  \label{tab:seg_IoU}
\end{table}

\subsubsection{Ablation Study}
\begin{table}[]
 \centering
 \caption{Results for ablation study. For each method we report class wise and mean IoU.}
 \vspace{-0.1cm}
 \begin{tabular}{|c|c|c|c|c|}
 \hline
 Method&Cars&Pedestrians&Bicyclists&meanIoU\\\hline
 100 Layer Tiramisu~\cite{jegou2017one}&74.2&\textbf{48.7}&43.7&55.5\\
 TD block~\cite{jegou2017one}&72.2&48.3&41.2&53.9\\
 Down-sample 8$\times$&74.1&43.8&39.7&52.5\\
 Down-sample width $4\times$&74.7&45.0&38.6&52.8\\ 
 db\_3 depth separable&74.2&49.2&36.8&53.4\\
 db\_3 + db\_2 depth separable&73.6&41.2&33.2&49.3\\
 DBLiDARNet&\textbf{75.1}&47.4&\textbf{45.4}&\textbf{56.0}\\
  \hline
 \end{tabular}
  \label{tab:seg_IoU_ablation_study}
\end{table}

In this ablation study, we justify the network design choices we mentioned in~\secref{sec:seg_nn}. The main differences between our dense blocks based fully convolutional network and the architecture proposed by~\citeauthor{jegou2017one}~\cite{jegou2017one}.
\begin{enumerate}
\item Replacing the transition-down block with max pooling for down-sampling the feature maps. This block implements a composite function comprising of batch normalization, ReLU activation, convolution layer ($1\times1$), dropout and max-pooling. We replace this transition-down block by a max-pooling layer. This decision is based on our empirical findings.
\item Using depth separable convolution layers instead of convolution layers for dense blocks in the decoder. This helps in reducing the parameters from~3.6M to~2.8M.      
\end{enumerate}
The architecture proposed by~\citeauthor{jegou2017one}~\cite{jegou2017one} consists of five transition down blocks for down-sampling the feature maps 32$\times$. They, therefore use five up-convolution layers in the decoder along with same number of dense blocks. Such a high down-sampling rate will result in significant loss of information for reasons discussed before (\secref{sec:seg_nn}). Therefore in our implementation of their architecture we only use two transition down blocks instead of five. In~\tabref{tab:seg_IoU_ablation_study} we report results for a model where we use our architecture but replace max-pool layers with transition down (TD) blocks. Our proposed architecture outperforms their architecture marginally while using fewer parameters. Using transition down blocks instead of a max-pooling layer leads to a slight decrease in performance as well.
Comparing different down-sampling strategies, we trained two different models. For the first model we down-sample 8$\times$ instead of 4 and for the second model we down-sample 4$\times$ but only along the width dimension while keeping the height unchanged, similar to~\cite{wu2018squeezeseg}. As reported in~\tabref{tab:seg_IoU_ablation_study}, for the first model (down-sample 8$\times$), the IoU for the class \textit{car} remains comparable but a decrease in performance is observed for the other classes. In comparison to cars, pedestrians and bicyclists are smaller and therefore a large down-sampling rate adversely affects these classes in comparison to other classes. For the second model, similar to the first, a noticeable decrease in performance is observed for both \textit{pedestrian} and \textit{bicyclist} classes. Without decreasing the height, feature maps have larger spatial resolution, thereby requiring more operations. Even though large down-sampling rate can hamper the performance, especially for the task of semantic segmentation, it is still required for increasing the receptive field as well as making the model efficient considering both the memory and computational requirements. Our proposed strategy of down-sampling the feature maps 4$\times$ allows us to exploit the advantages of such operations without losing the crucial information necessary for predicting accurate segmentation masks. 

We trained two models, where we use depth separable convolution only in the last dense block of the encoder (db\_3) and then in last two dense blocks together (db\_3 + db\_2). In both cases performance decreases, especially for the second case the decrease is substantial. Even though depth separable convolution is an ingenious way of reducing parameters but excessively using it can decrease performance as well.
\subsubsection{Semantic KITTI}
In \tabref{tab:results_segmentation_semantic_kitti}, we report results for the semantic KITTI dataset~\cite{behley2019iccv}.
\setlength\tabcolsep{3.5pt}\begin{table*}[ht]\centering\footnotesize{
\begin{tabular}{lc|ccccccccccccccccccc}
\toprule 
Approach & \begin{sideways}mIoU\end{sideways} & \begin{sideways}road\end{sideways} & \begin{sideways}sidewalk\end{sideways} & \begin{sideways}parking\end{sideways} & \begin{sideways}other-ground\end{sideways} & \begin{sideways}building\end{sideways} & \begin{sideways}car\end{sideways} & \begin{sideways}truck\end{sideways} & \begin{sideways}bicycle\end{sideways} & \begin{sideways}motorcycle\end{sideways} & \begin{sideways}other-vehicle\end{sideways} & \begin{sideways}vegetation\end{sideways} & \begin{sideways}trunk\end{sideways} & \begin{sideways}terrain\end{sideways} & \begin{sideways}person\end{sideways} & \begin{sideways}bicyclist\end{sideways} & \begin{sideways}motorcyclist\end{sideways} & \begin{sideways}fence\end{sideways} & \begin{sideways}pole\end{sideways} & \begin{sideways}traffic sign\end{sideways}\\
\midrule
PointNet \cite{qi2017pointnet}  & 14.6 & 61.6 & 35.7 & 15.8 & 1.4 & 41.4 & 46.3 & 0.1 & 1.3 & 0.3 & 0.8 & 31.0 & 4.6 & 17.6 & 0.2 & 0.2 & 0.0 & 12.9 & 2.4 & 3.7\\
SPGraph \cite{landrieu2018cvpr}  & 17.4 & 45.0 & 28.5 & 0.6 & 0.6 & 64.3 & 49.3 & 0.1 & 0.2 & 0.2 & 0.8 & 48.9 & 27.2 & 24.6 & 0.3 & 2.7 & 0.1 & 20.8 & 15.9 & 0.8\\
SPLATNet \cite{su2018cvpr}  & 18.4 & 64.6 & 39.1 & 0.4 & 0.0 & 58.3 & 58.2 & 0.0 & 0.0 & 0.0 & 0.0 & 71.1 & 9.9 & 19.3 & 0.0 & 0.0 & 0.0 & 23.1 & 5.6 & 0.0\\
PointNet++ \cite{qi2017pointnet++}  & 20.1 & 72.0 & 41.8 & 18.7 & 5.6 & 62.3 & 53.7 & 0.9 & 1.9 & 0.2 & 0.2 & 46.5 & 13.8 & 30.0 & 0.9 & 1.0 & 0.0 & 16.9 & 6.0 & 8.9\\
SqueezeSeg \cite{wu2018squeezeseg}  & 29.5 & 85.4 & 54.3 & 26.9 & 4.5 & 57.4 & 68.8 & 3.3 & 16.0 & 4.1 & 3.6 & 60.0 & 24.3 & 53.7 & 12.9 & 13.1 & 0.9 & 29.0 & 17.5 & 24.5\\
SqueezeSegV2 \cite{wu2018squeezesegv2}  & 39.7 & 88.6 & 67.6 & 45.8 & 17.7 & 73.7 & 81.8 & 13.4 & 18.5 & 17.9 & 14.0 & 71.8 & 35.8 & 60.2 & 20.1 & 25.1 & 3.9 & 41.1 & 20.2 & 36.3\\
TangentConv \cite{tatarchenko2018cvpr}  & 40.9 & 83.9 & 63.9 & 33.4 & 15.4 & 83.4 & 90.8 & 15.2 & 2.7 & 16.5 & 12.1 & 79.5 & 49.3 & 58.1 & 23.0 & 28.4 & 8.1 & 49.0 & 35.8 & 28.5\\
DarkNet21Seg  & 47.4 & 91.4 & 74.0 & 57.0 & 26.4 & 81.9 & 85.4 & 18.6 & 26.2 & 26.5 & 15.6 & 77.6 & 48.4 & 63.6 & 31.8 & 33.6 & 4.0 & 52.3 & 36.0 & 50.0\\
DarkNet53Seg  & 49.9 & 91.8 & 74.6 & 64.8 & 27.9 & 84.1 & 86.4 & 25.5 & 24.5 & 32.7 & 22.6 & 78.3 & 50.1 & 64.0 & 36.2 & 33.6 & 4.7 & 55.0 & 38.9 & 52.2\\
DBLiDARNet&37.6&85.8&59.3&8.7&1.0&78.6&81.5&6.6&29.4&19.6&6.5&77.1&46.0&58.1&23.7&20.1&2.4&39.6&32.6&39.1\\
\bottomrule
\end{tabular}}
\setlength\tabcolsep{6.0pt}\caption{Single scan results (19 classes) for all baselines on sequences 11 to 21 (test set). All methods were trained on sequences 00 to 10, except for sequence 08 which is used as validation set.}
\label{tab:results_segmentation_semantic_kitti}
\end{table*}

\subsection{Bayes Filter}
\label{sec:results_object_bayes_filter}
To evaluate our proposed Bayes filter approach, we use the KITTI tracking benchmark. The benchmark contains 20 sequences and to evaluate our approach on all of the sequences, we split the sequences into two different sets. We train our network on both sets separately and use the other set for testing i.e. we train a model on the first set and test the learned model on the second set and then train on the second set and test on the first set. 


For training the network we use our proposed network with the exact same parameters as discussed in~\secref{sec:seg_training}, with the one difference. In this case the input resolution of the images are $64 \times 324 \times 5$, in comparison to $64 \times 512 \times 5$. For evaluating the proposed Bayes filter we use our network as the baseline method and report comparison with the segmentation results from the network. In~\figref{fig:object_bayes_filter}, we illustrate the differences in the segmentation results for a sequence of six consecutive scans. 
In the case of neural network, points on a car are correctly classified in the first scan but in the next few scans, points on the same car are misclassified as background. For the same scans, our proposed Bayes filter is able to consistently classify points on the car correctly.
\begin{figure*}[]
     \centering
 	  \begin{subfigure}[]{0.3\textwidth}
      \includegraphics[width=0.75\linewidth]{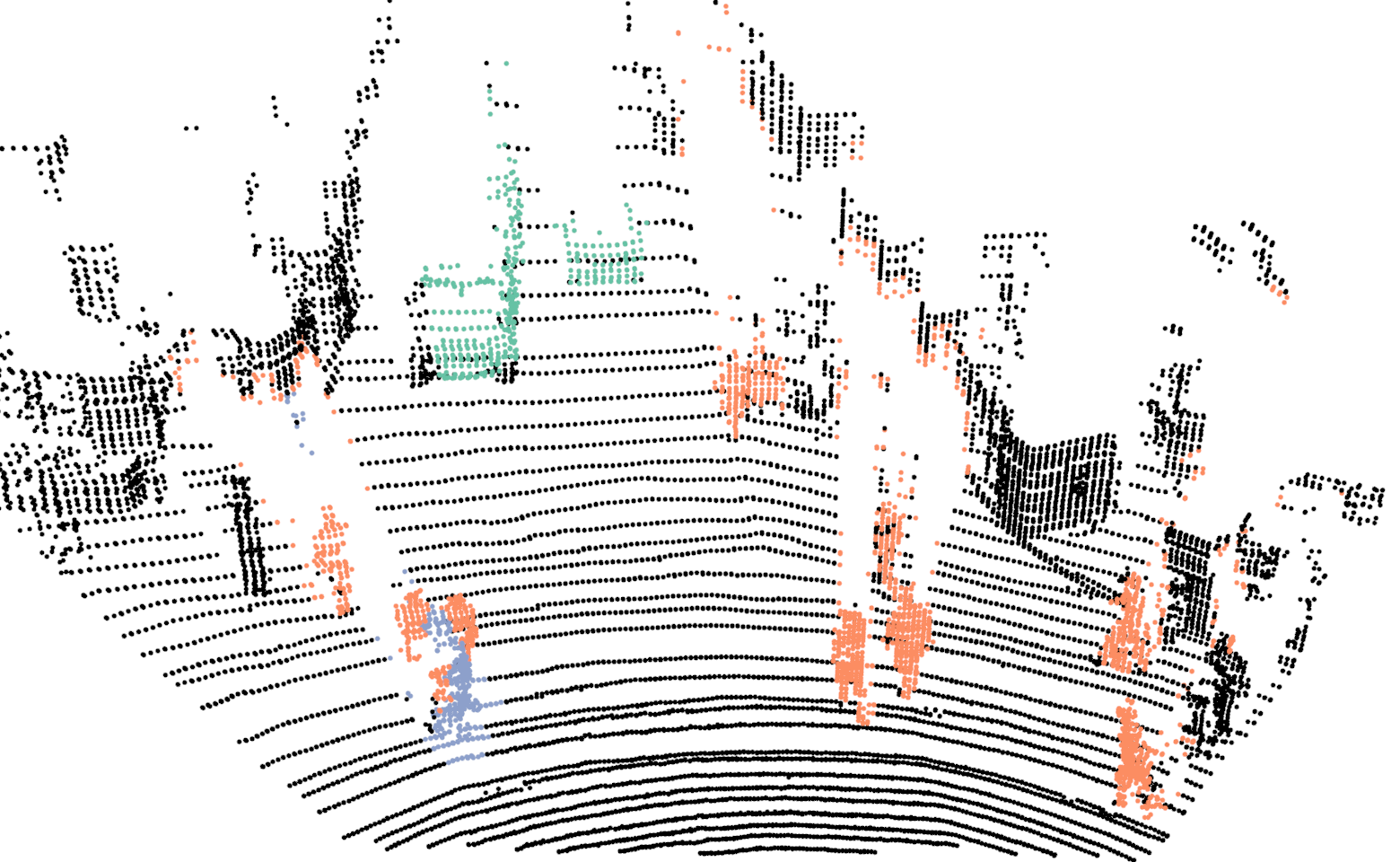}
    \caption{}
    \label{fig:resnet_corr}
  \end{subfigure}%
  \centering
  \begin{subfigure}[]{0.3\textwidth}
    \centering
    \includegraphics[width=0.65\linewidth]{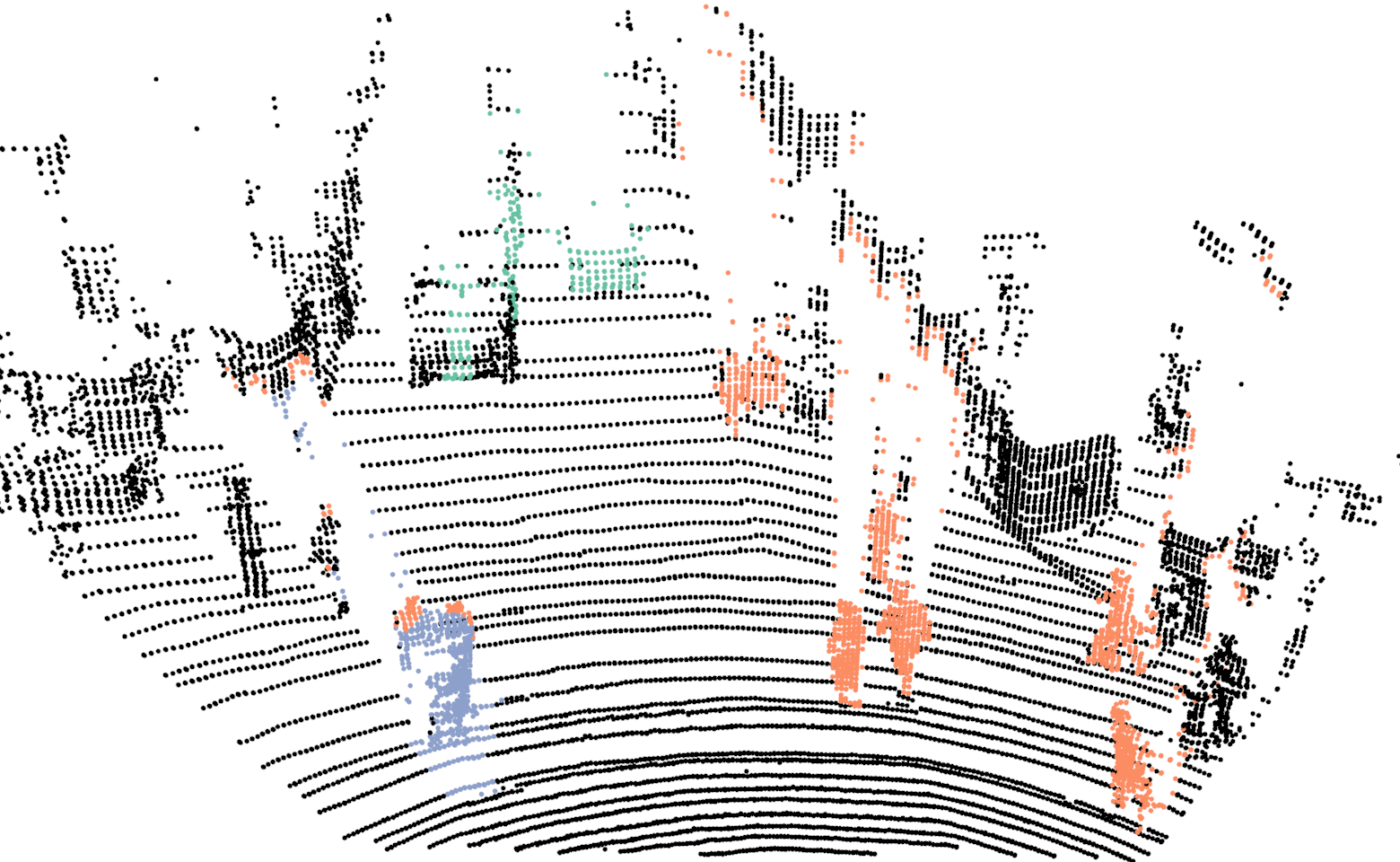}
    \caption{}
    \label{fig:resnet_tsne}
    \end{subfigure}%
      \centering
 	  \begin{subfigure}[]{0.3\textwidth}
      \includegraphics[width=0.65\linewidth]{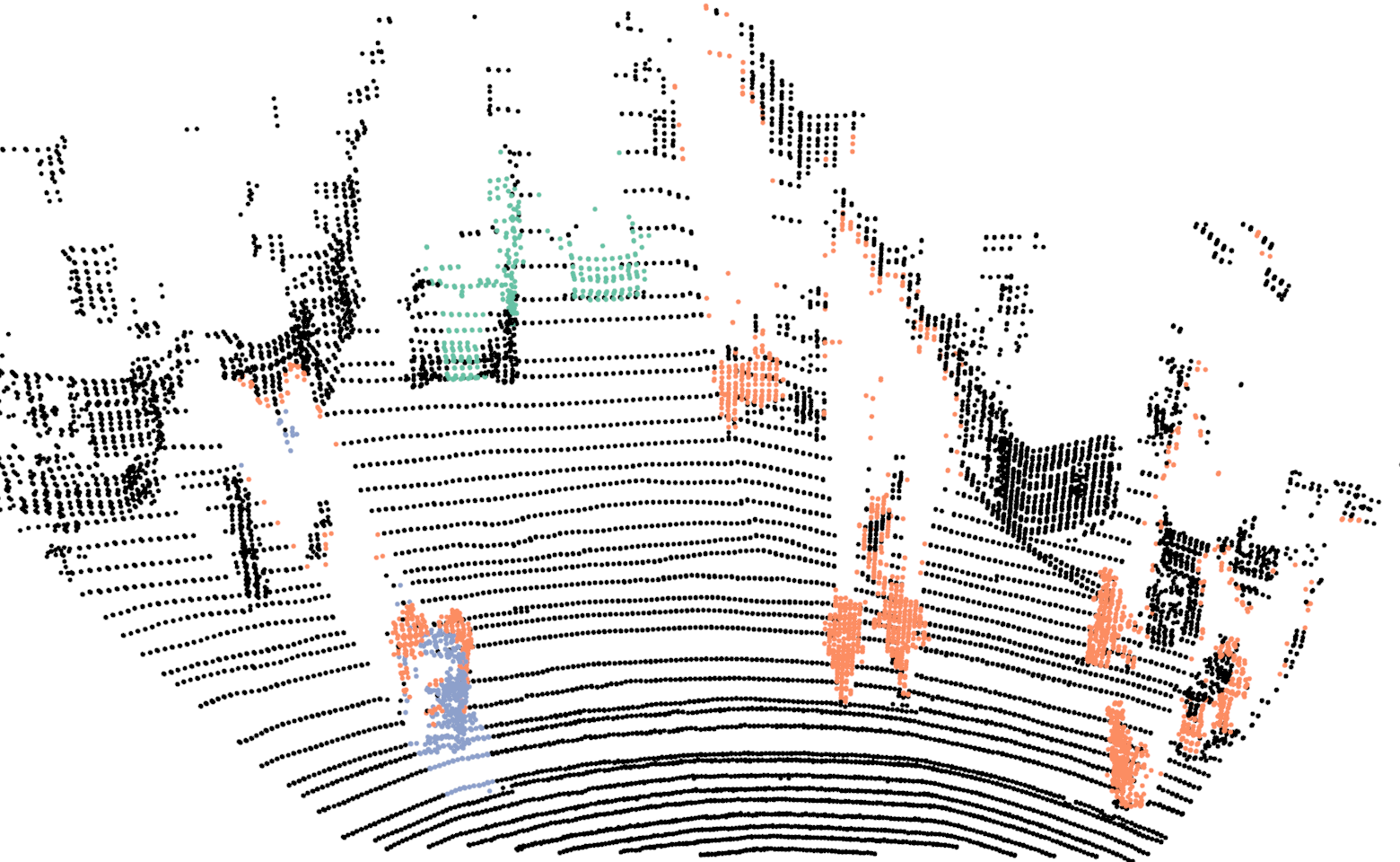}
    \caption{}
    \label{fig:matchnet_corr}
  \end{subfigure}%
  \\
       \centering
 	  \begin{subfigure}[]{0.3\textwidth}
      \includegraphics[width=0.65\linewidth]{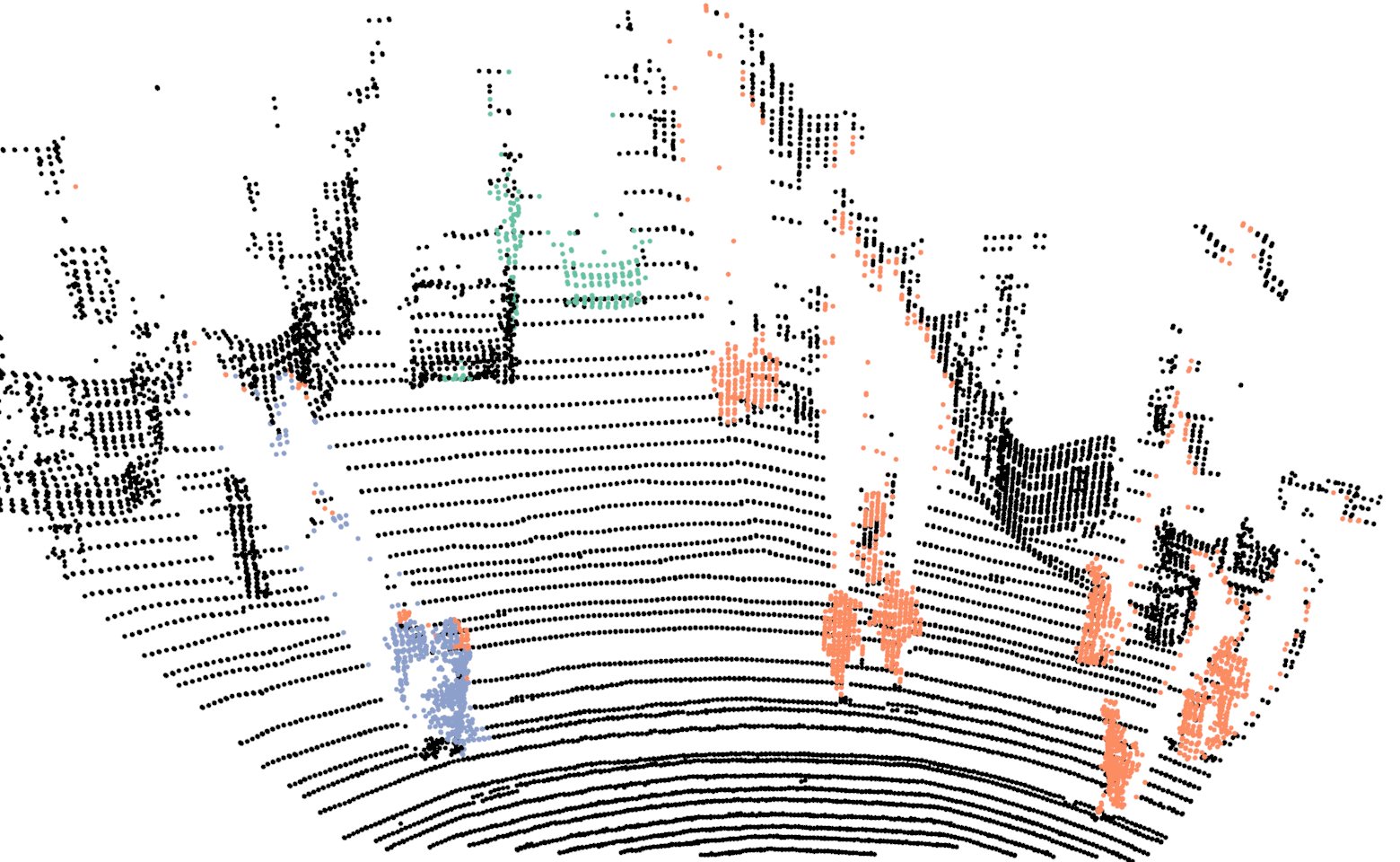}
    \caption{}
    \label{fig:resnet_corr}
  \end{subfigure}%
  \centering
  \begin{subfigure}[]{0.3\textwidth}
    \centering
    \includegraphics[width=0.65\linewidth]{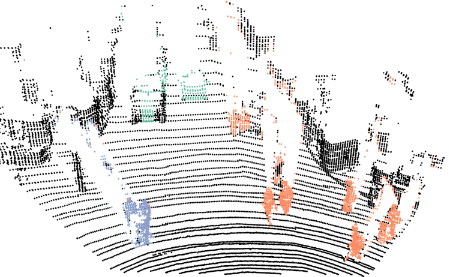}
    \caption{}
    \label{fig:resnet_tsne}
    \end{subfigure}%
      \centering
 	  \begin{subfigure}[]{0.3\textwidth}
      \includegraphics[width=0.65\linewidth]{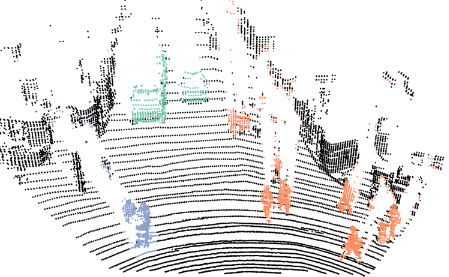}
    \caption{}
    \label{fig:matchnet_corr}
  \end{subfigure}%
  \\
       \centering
 	  \begin{subfigure}[]{0.3\textwidth}
      \includegraphics[width=0.65\linewidth]{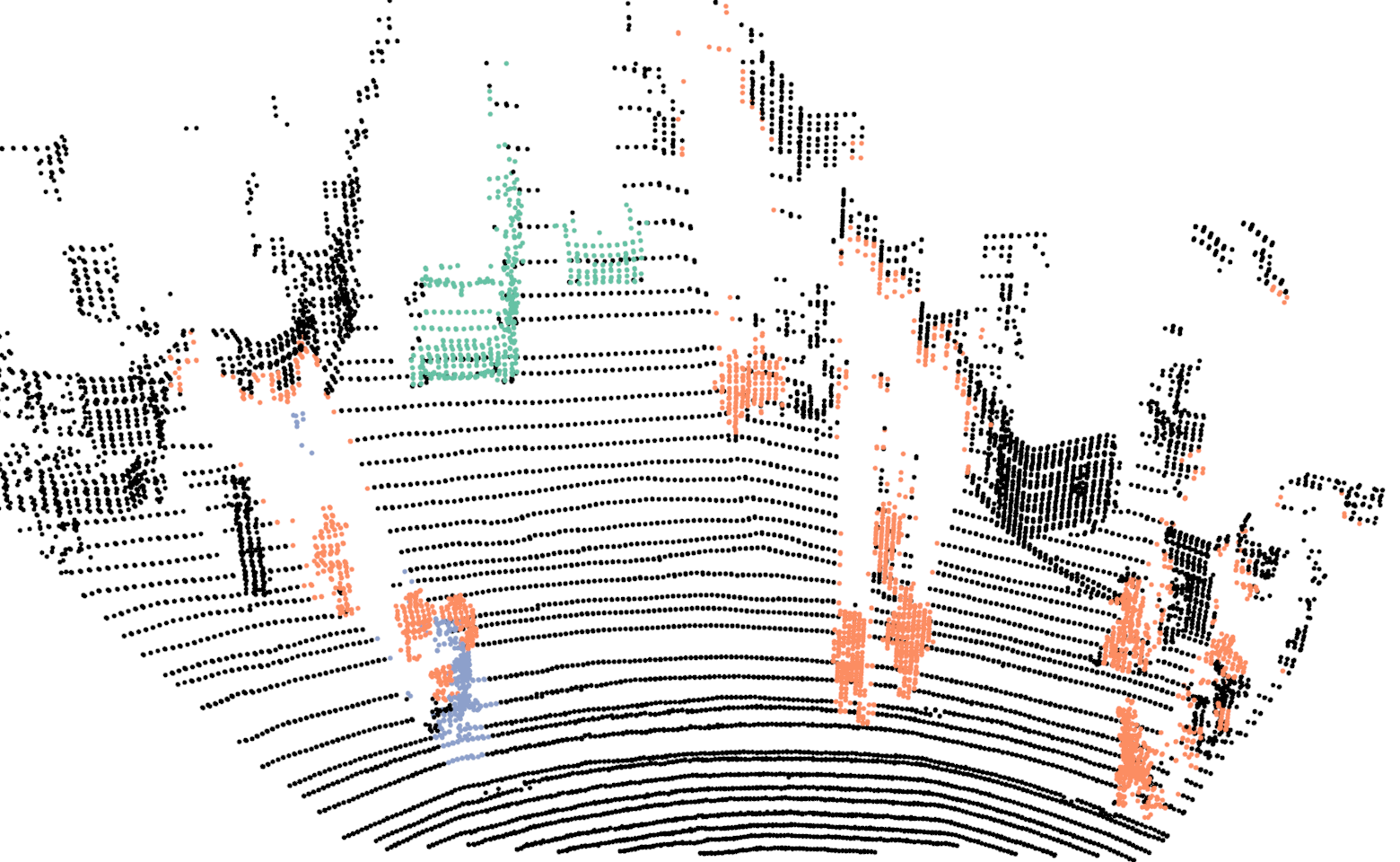}
    \caption{}
    \label{fig:resnet_corr}
  \end{subfigure}%
  \centering
  \begin{subfigure}[]{0.3\textwidth}
    \centering
    \includegraphics[width=0.65\linewidth]{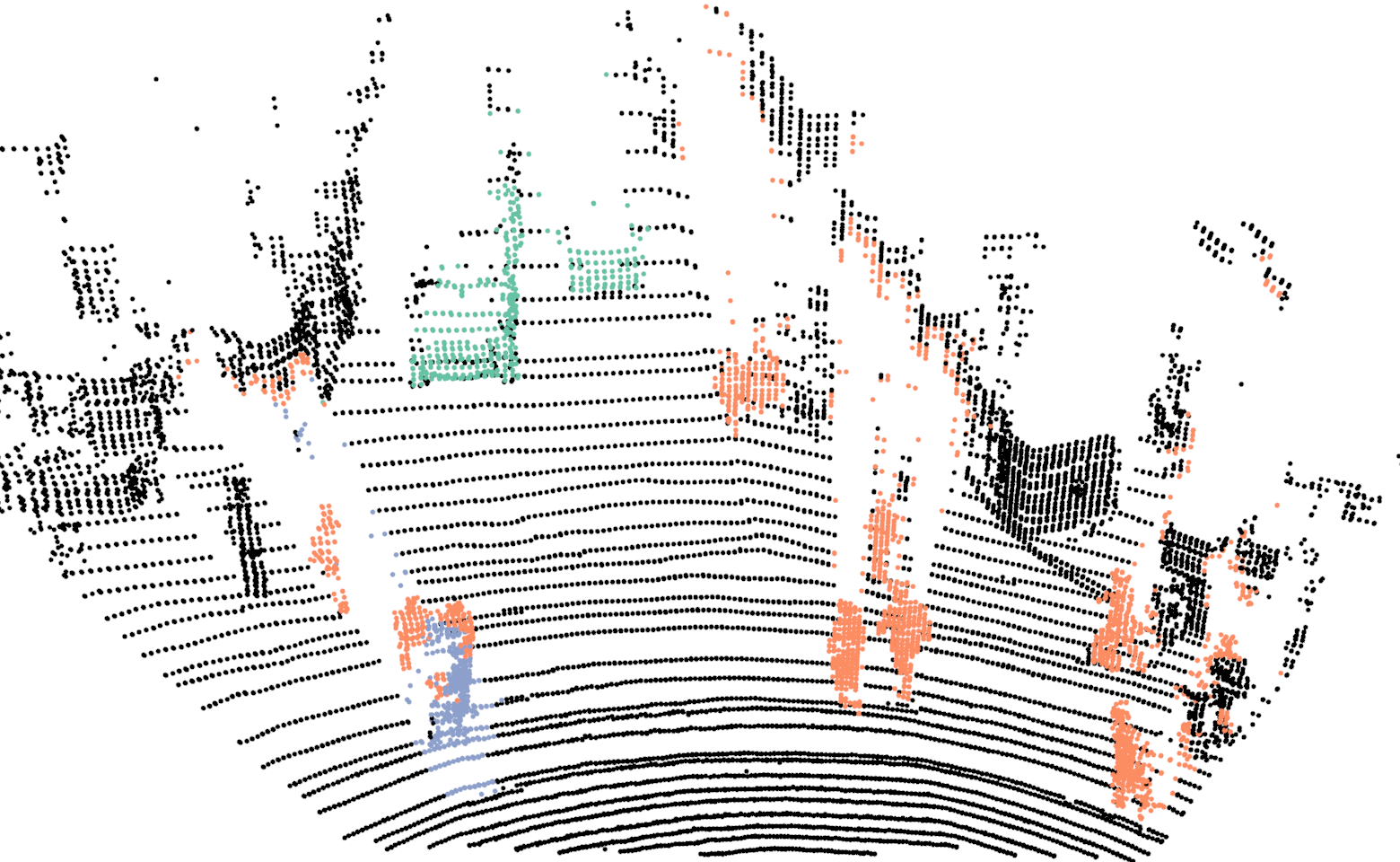}
    \caption{}
    \label{fig:resnet_tsne}
    \end{subfigure}%
      \centering
 	  \begin{subfigure}[]{0.3\textwidth}
      \includegraphics[width=0.65\linewidth]{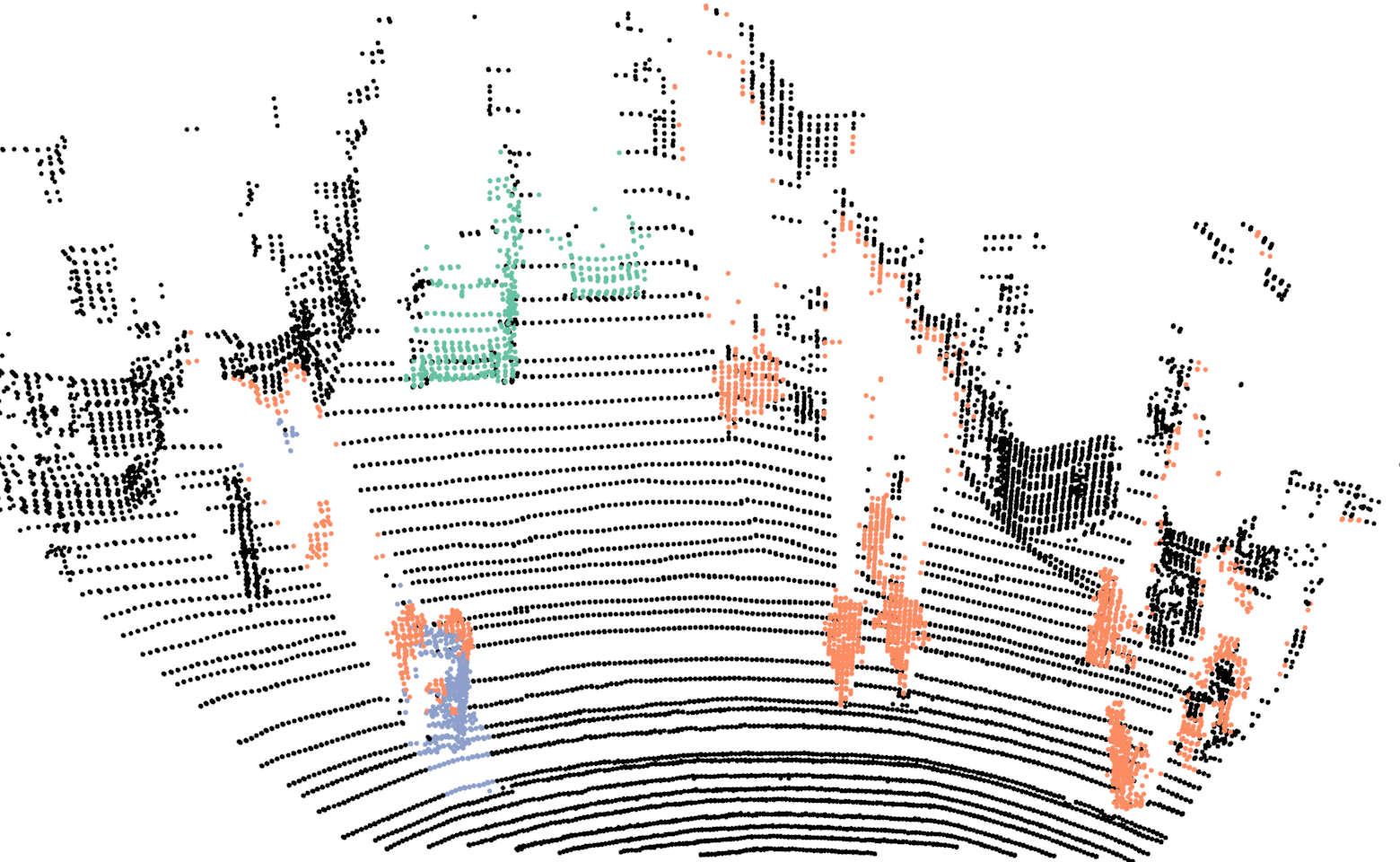}
    \caption{}
    \label{fig:matchnet_corr}
  \end{subfigure}%
  \\
       \centering
 	  \begin{subfigure}[]{0.3\textwidth}
      \includegraphics[width=0.65\linewidth]{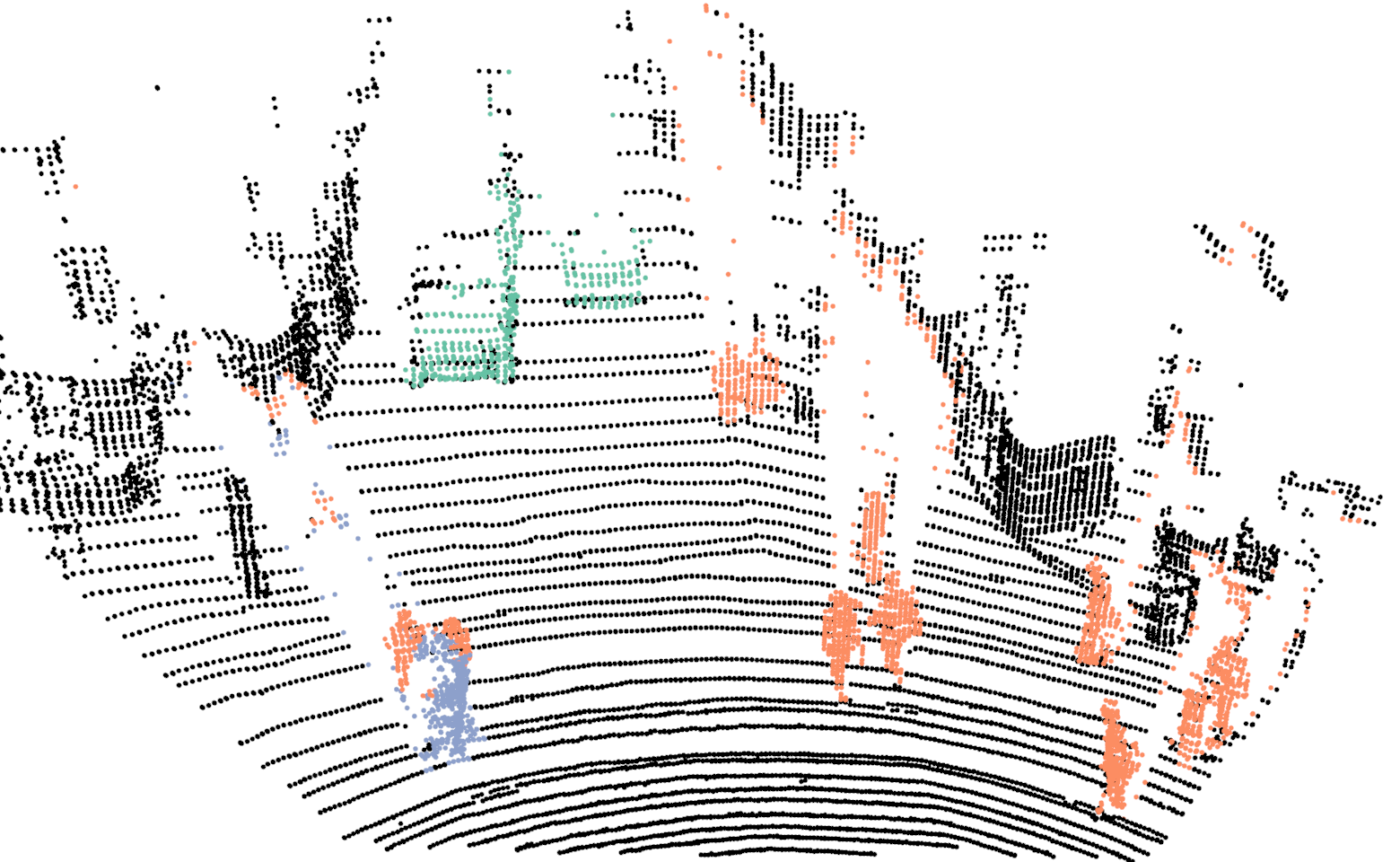}
    \caption{}
    \label{fig:resnet_corr}
  \end{subfigure}%
  \centering
  \begin{subfigure}[]{0.3\textwidth}
    \centering
    \includegraphics[width=0.65\linewidth]{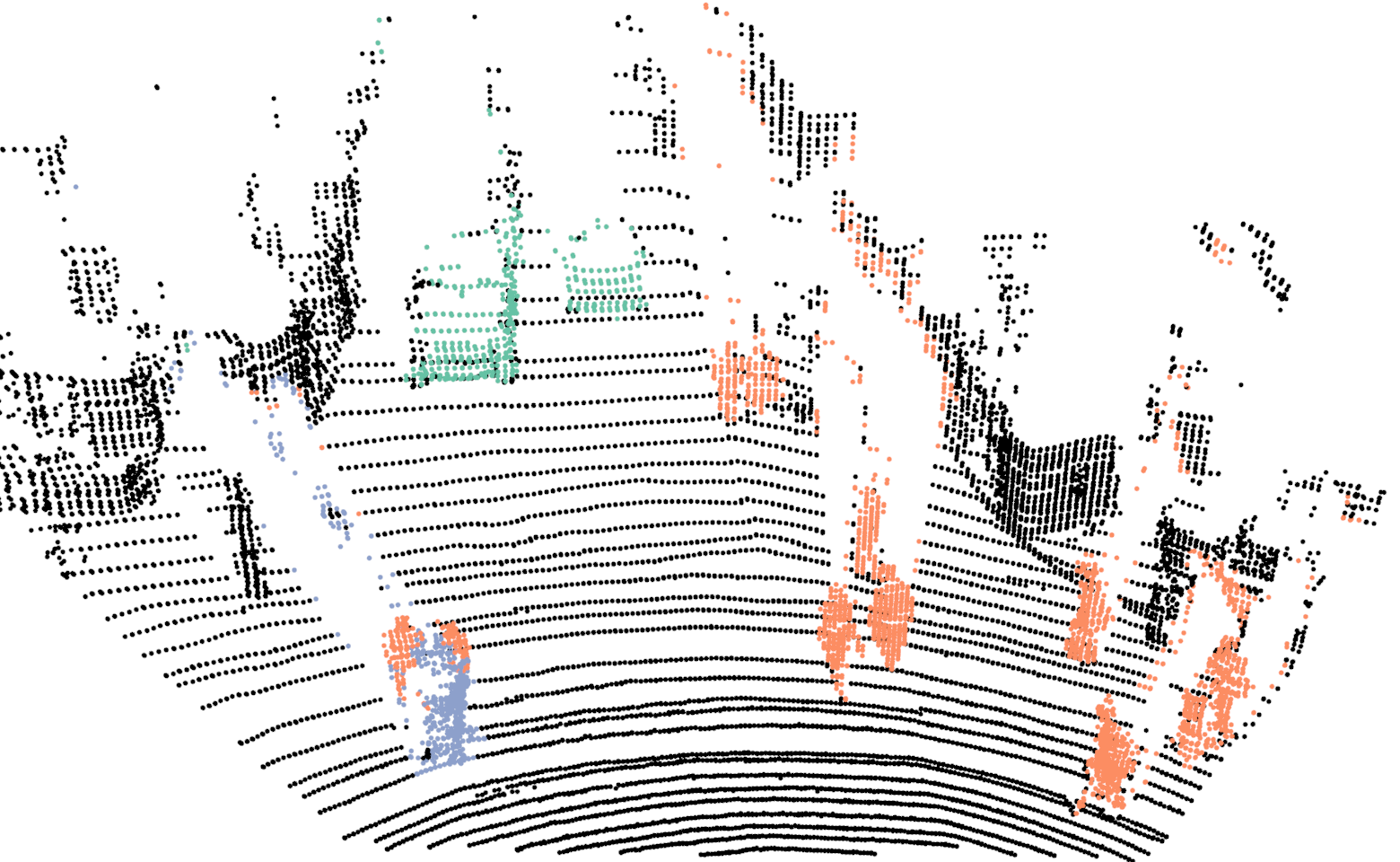}
    \caption{}
    \label{fig:resnet_tsne}
    \end{subfigure}%
      \centering
 	  \begin{subfigure}[]{0.3\textwidth}
      \includegraphics[width=0.65\linewidth]{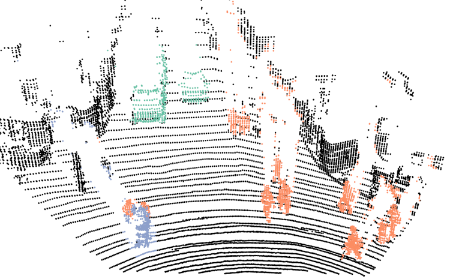}
    \caption{}
    \label{fig:matchnet_corr}
  \end{subfigure}%
    \caption[Illustration of semantic segmentation with the object Bayes filter]{Illustration of semantic segmentation with the object Bayes filter. In the top two rows ((a)-(f)), we show the output of our proposed DCNN, for six consecutive scans. In the top left image, points on a car (top left) are correctly classified, but in subsequent scans, points on the same car are first partially ((b)-(c)) and then completely ((d)) misclassified as background. In the bottom two rows, we show the output of our proposed binary Bayes filter for the same six consecutive scans. For all the six scans, points on the same car are correctly classified. These results clearly illustrate that our proposed Bayes filter method is able to successfully mitigate the sporadic erroneous predictions from the neural network.}
\label{fig:object_bayes_filter}    
\end{figure*}   

In~\tabref{tab:object_bayes_filter_results}, we report class wise IoU for different sequences, for both our DCNN and the Bayes filter approach. In the cases where no instances of a class is observed, we do not report results as well (indicated by a dash sign). The performance of our DCNN on this dataset is similar to the results reported in ~\secref{sec:segmentation_results}. For some sequences, the IoU for the class bicyclists is zero. In these cases, majority of times these objects are either far from the sensor or occluded and in the rare cases they are misclassified as pedestrians. 


Comparing the DCNN results with the Bayes filter approach, across different sequences and classes, an improvement in IoU is consistently observed after using the Bayes filter approach. For most cases the improvement in IoU is around 4\% to 9\% but an improvement of 27\% is achieved for class \textit{pedestrian} in sequence 2 and staggering improvement of 51\% is achieved for class \textit{bicyclist} in sequence 4. For couple of isolated cases, a decrease in IoU is observed after using the filter approach. The implicit assumption of our Bayes filter approach is that the predictions from DCNN is seldom wrong and for cases, the filter uses the previous knowledge to correct those predictions. In the rare cases where this assumption is violated, the information accumulated by the filter spurs from incorrect measurements and therefore the filter approach needs multiple correct predictions from DCNN to improve its knowledge in comparison to a single prediction needed by DCNN. For instance, in the sequence 0, points on a bicyclist were labeled as pedestrian more than often, causing Bayes filter to accumulate the incorrect predictions. 


\begin{table}[]
 \centering
 \caption{Class wise IoU for DCNN and the binary object Bayes filter}
 \vspace{-0.1cm}
 \begin{tabular}{|c|c|c|c|c|c|c|}
 \hline
 \multirow{2}{*}{Seq. ID}&\multicolumn{3}{ c|}{DBLiDARNet}
 &\multicolumn{3}{ c |}{Object Bayes Filter}\\\cline{2-7}
 &Cars&Pedestrians&Bicyclist&Cars&Pedestrians&Bicyclist\\
 \hline
 0&76.2&2.0&\textbf{29.6}&\textbf{79.2}&2.0&23.6\\
 2&54.9&37.0&0.0&\textbf{55.3}&\textbf{46.9}&0.0\\
 3&75.2&-&-&\textbf{75.5}&-&-\\
 4&66.6&40.8&35.2&\textbf{69.1}&\textbf{47.4}&\textbf{53.2}\\
 5&\textbf{70.1}&-&-&70.0&-\\
 6&\textbf{87.2}&-&-&87.1&-&-\\
 7&83.2&28.2&-&\textbf{83.5}&\textbf{32.7}&-\\
 8&66.9&-&-&\textbf{69.9}&-&-\\
 9&71.9&18.6&-&\textbf{72.9}&\textbf{21.6}&-\\
 10&72.4&0.0&0.0&\textbf{75.1}&0.0&0.0\\
 11&88.4&\textbf{15.6}&-&\textbf{89.6}&15.3&-\\
 12&51.5&0.0&\textbf{4.0}&\textbf{58.5}&0.0&1.6\\
 13&24.2&\textbf{50.7}&39.5&\textbf{31.3}&50.6&\textbf{41.1}\\
 14&\textbf{89.6}&40.2&-&86.3&42.6&-\\
 15&83.9&70.1&5.0&\textbf{85.7}&\textbf{72.5}&5.0\\
 16&63.8&75.3&54.5&\textbf{64.1}&\textbf{77.0}&\textbf{60.7}\\
 17&-&81.8&0.0&-&\textbf{83.7}&0.0\\
 18&84.7&-&-&84.7&-&-\\
 19&68.4&66.2&36.9&\textbf{74.0}&\textbf{66.1}&\textbf{37.8}\\
 20&69.1&-&-&\textbf{69.4}&-&-\\
   \hline
  \end{tabular}
  \label{tab:object_bayes_filter_results}
\end{table}

\section{Conclusions}
In this paper, we proposed a DCNN to segment points in a 3D LiDAR scan into multiple semantic categories. Our proposed architecture is based on dense blocks and uses depth separable convolution to reduce the parameters while still maintaining competitive performance. It significantly outperforms state-of-the-art neural network architectures, with an average improvement of around 16\% across different classes. In the presented ablation study, we justify our architecture choices. The neural network predicts the segmentation mask for each scan independently and to make these predictions temporally consistent, we proposed a Bayes filter method. Through extensive evaluation on the KITTI tracking benchmark, we report a consistent improvement across classes and sequences. 


\bibliographystyle{plainnat}
\footnotesize
\bibliography{dewan20icra}

\begin{thebibliography}{28}
\providecommand{\natexlab}[1]{#1}
\providecommand{\url}[1]{\texttt{#1}}
\expandafter\ifx\csname urlstyle\endcsname\relax
  \providecommand{\doi}[1]{doi: #1}\else
  \providecommand{\doi}{doi: \begingroup \urlstyle{rm}\Url}\fi

\bibitem[way(2019)]{waymo-mission}
{Waymo: Our Mission}.
\newblock \url{https://waymo.com/mission/}, 2019.
\newblock [Online; accessed 28-May-2019].

\bibitem[Abadi et~al.(2016)Abadi, Agarwal, Barham, Brevdo, Chen, Citro,
  Corrado, Davis, Dean, Devin, et~al.]{abadi2016tensorflow}
Mart{\'\i}n Abadi, Ashish Agarwal, Paul Barham, Eugene Brevdo, Zhifeng Chen,
  Craig Citro, Greg~S Corrado, Andy Davis, Jeffrey Dean, Matthieu Devin, et~al.
\newblock Tensorflow: Large-scale machine learning on heterogeneous distributed
  systems.
\newblock \emph{arXiv preprint arXiv:1603.04467}, 2016.

\bibitem[Badrinarayanan et~al.(2015)Badrinarayanan, Kendall, and
  Cipolla]{kendall2015}
Vijay Badrinarayanan, Alex Kendall, and Roberto Cipolla.
\newblock Segnet: {A} deep convolutional encoder-decoder architecture for image
  segmentation.
\newblock \emph{arXiv preprint arXiv: 1511.00561}, 2015.
\newblock URL \url{http://arxiv.org/abs/1511.00561}.

\bibitem[Behley et~al.(2019)Behley, Garbade, Milioto, Quenzel, Behnke,
  Stachniss, and Gall]{behley2019iccv}
J.~Behley, M.~Garbade, A.~Milioto, J.~Quenzel, S.~Behnke, C.~Stachniss, and
  J.~Gall.
\newblock {SemanticKITTI: A Dataset for Semantic Scene Understanding of LiDAR
  Sequences}.
\newblock In \emph{Proc. of the IEEE/CVF International Conf.~on Computer Vision
  (ICCV)}, 2019.

\bibitem[Chen et~al.(2017)Chen, Papandreou, Schroff, and
  Adam]{chen2017rethinking}
Liang-Chieh Chen, George Papandreou, Florian Schroff, and Hartwig Adam.
\newblock Rethinking atrous convolution for semantic image segmentation.
\newblock \emph{arXiv preprint arXiv:1706.05587}, 2017.

\bibitem[Chen et~al.(2018)Chen, Papandreou, Kokkinos, Murphy, and
  Yuille]{chen2018deeplab}
Liang-Chieh Chen, George Papandreou, Iasonas Kokkinos, Kevin Murphy, and Alan~L
  Yuille.
\newblock Deeplab: Semantic image segmentation with deep convolutional nets,
  atrous convolution, and fully connected crfs.
\newblock \emph{IEEE transactions on pattern analysis and machine
  intelligence}, 40\penalty0 (4):\penalty0 834--848, 2018.

\bibitem[Chollet(2017)]{chollet2017xception}
Fran{\c{c}}ois Chollet.
\newblock Xception: Deep learning with depthwise separable convolutions.
\newblock In \emph{Proceedings of the IEEE conference on computer vision and
  pattern recognition}, pages 1251--1258, 2017.

\bibitem[Dewan et~al.(2016)Dewan, Caselitz, Tipaldi, and
  Burgard]{dewan2016rigid}
Ayush Dewan, Tim Caselitz, Gian~Diego Tipaldi, and Wolfram Burgard.
\newblock Rigid scene flow for 3d lidar scans.
\newblock In \emph{IEEE/RSJ International Conference on Intelligent Robots and
  Systems (IROS)}, 2016.

\bibitem[Dewan et~al.(2017)Dewan, Oliveira, and Burgard]{dewan2017deep}
Ayush Dewan, Gabriel~L Oliveira, and Wolfram Burgard.
\newblock Deep semantic classification for 3d lidar data.
\newblock In \emph{2017 IEEE/RSJ International Conference on Intelligent Robots
  and Systems (IROS)}, pages 3544--3549. IEEE, 2017.

\bibitem[Hu et~al.(2018)Hu, Shen, and Sun]{hu2018squeeze}
Jie Hu, Li~Shen, and Gang Sun.
\newblock Squeeze-and-excitation networks.
\newblock In \emph{Proceedings of the IEEE conference on computer vision and
  pattern recognition}, pages 7132--7141, 2018.

\bibitem[Huang et~al.(2016)Huang, Liu, Weinberger, and van~der
  Maaten]{huang2016densely}
Gao Huang, Zhuang Liu, Kilian~Q Weinberger, and Laurens van~der Maaten.
\newblock Densely connected convolutional networks.
\newblock \emph{arXiv preprint arXiv:1608.06993}, 2016.

\bibitem[Iandola et~al.(2016)Iandola, Han, Moskewicz, Ashraf, Dally, and
  Keutzer]{iandola2016squeezenet}
Forrest~N Iandola, Song Han, Matthew~W Moskewicz, Khalid Ashraf, William~J
  Dally, and Kurt Keutzer.
\newblock Squeezenet: Alexnet-level accuracy with 50x fewer parameters and< 0.5
  mb model size.
\newblock \emph{arXiv preprint arXiv:1602.07360}, 2016.

\bibitem[J{\'e}gou et~al.(2017)J{\'e}gou, Drozdzal, Vazquez, Romero, and
  Bengio]{jegou2017one}
Simon J{\'e}gou, Michal Drozdzal, David Vazquez, Adriana Romero, and Yoshua
  Bengio.
\newblock The one hundred layers tiramisu: Fully convolutional densenets for
  semantic segmentation.
\newblock In \emph{Computer Vision and Pattern Recognition Workshops (CVPRW),
  2017 IEEE Conference on}, pages 1175--1183. IEEE, 2017.

\bibitem[Kingma and Ba(2014)]{kingma2014adam}
Diederik Kingma and Jimmy Ba.
\newblock Adam: A method for stochastic optimization.
\newblock \emph{arXiv preprint arXiv:1412.6980}, 2014.

\bibitem[Landrieu and Simonovsky(2018)]{landrieu2018cvpr}
Loic Landrieu and Martin Simonovsky.
\newblock {Large-scale Point Cloud Semantic Segmentation with Superpoint
  Graphs}.
\newblock 2018.
\newblock URL \url{https://arxiv.org/pdf/1711.09869}.

\bibitem[Lin et~al.(2017)Lin, Goyal, Girshick, He, and
  Doll{\'a}r]{lin2017focal}
Tsung-Yi Lin, Priya Goyal, Ross Girshick, Kaiming He, and Piotr Doll{\'a}r.
\newblock Focal loss for dense object detection.
\newblock In \emph{Proceedings of the IEEE international conference on computer
  vision}, pages 2980--2988, 2017.

\bibitem[Long et~al.(2015)Long, Shelhamer, and Darrell]{long_shelhamer_fcn}
Jonathan Long, Evan Shelhamer, and Trevor Darrell.
\newblock Fully convolutional networks for semantic segmentation.
\newblock \emph{IEEE Conference on Computer Vision and Pattern Recognition
  (CVPR)}, 2015.

\bibitem[Naseer et~al.(2017)Naseer, Oliveira, Brox, and
  Burgard]{naseer2017semantics}
Tayyab Naseer, Gabriel~L Oliveira, Thomas Brox, and Wolfram Burgard.
\newblock Semantics-aware visual localization under challenging perceptual
  conditions.
\newblock In \emph{2017 IEEE International Conference on Robotics and
  Automation (ICRA)}, pages 2614--2620. IEEE, 2017.

\bibitem[Qi et~al.(2017{\natexlab{a}})Qi, Su, Mo, and Guibas]{qi2017pointnet}
Charles~R Qi, Hao Su, Kaichun Mo, and Leonidas~J Guibas.
\newblock Pointnet: Deep learning on point sets for 3d classification and
  segmentation.
\newblock In \emph{Proceedings of the IEEE Conference on Computer Vision and
  Pattern Recognition}, pages 652--660, 2017{\natexlab{a}}.

\bibitem[Qi et~al.(2017{\natexlab{b}})Qi, Yi, Su, and Guibas]{qi2017pointnet++}
Charles~Ruizhongtai Qi, Li~Yi, Hao Su, and Leonidas~J Guibas.
\newblock Pointnet++: Deep hierarchical feature learning on point sets in a
  metric space.
\newblock In \emph{Advances in Neural Information Processing Systems}, pages
  5099--5108, 2017{\natexlab{b}}.

\bibitem[Radwan et~al.(2018)Radwan, Valada, and Burgard]{radwan2018vlocnet++}
Noha Radwan, Abhinav Valada, and Wolfram Burgard.
\newblock Vlocnet++: Deep multitask learning for semantic visual localization
  and odometry.
\newblock \emph{IEEE Robotics and Automation Letters}, 3\penalty0 (4):\penalty0
  4407--4414, 2018.

\bibitem[Ronneberger et~al.(2015)Ronneberger, Fischer, and
  Brox]{ronneberger2015u}
Olaf Ronneberger, Philipp Fischer, and Thomas Brox.
\newblock U-net: Convolutional networks for biomedical image segmentation.
\newblock In \emph{International Conference on Medical image computing and
  computer-assisted intervention}, pages 234--241. Springer, 2015.

\bibitem[Ruchti and Burgard(2018)]{ruchti18icra}
Philipp Ruchti and Wolfram Burgard.
\newblock Mapping with dynamic-object probabilities calculated from single 3d
  range scans.
\newblock In \emph{Proc.~of the IEEE Int. Conf. on Robotics \& Automation
  (ICRA)}, Brisbane, Australia, May 2018.
\newblock URL
  \url{http://ais.informatik.uni-freiburg.de/publications/papers/ruchti18icra.pdf}.

\bibitem[Su et~al.(2018)Su, Jampani, Sun, Maji, Kalogerakis, Yang, and
  Kautz]{su2018cvpr}
Hang Su, Varun Jampani, Deqing Sun, Subhransu Maji, Evangelos Kalogerakis,
  Ming-Hsuan Yang, and Jan Kautz.
\newblock {SPLATNet: Sparse Lattice Networks for Point Cloud Processing}.
\newblock 2018.
\newblock URL
  \url{http://openaccess.thecvf.com/content_cvpr_2018/html/../../content_cvpr_2018/papers/Su_SPLATNet_Sparse_Lattice_CVPR_2018_paper.pdf}.

\bibitem[Tatarchenko et~al.(2018)Tatarchenko, Park, Koltun, and
  Zhou]{tatarchenko2018cvpr}
Maxim Tatarchenko, Jaesik Park, Vladen Koltun, and Qian-Yi Zhou.
\newblock {Tangent Convolutions for Dense Prediction in 3D}.
\newblock 2018.
\newblock URL \url{proceedings: tatarchenko2018cvpr.pdf}.

\bibitem[Wang et~al.(2018)Wang, Shi, Yun, Tai, and Liu]{wang2018pointseg}
Yuan Wang, Tianyue Shi, Peng Yun, Lei Tai, and Ming Liu.
\newblock Pointseg: Real-time semantic segmentation based on 3d lidar point
  cloud.
\newblock \emph{arXiv preprint arXiv:1807.06288}, 2018.

\bibitem[Wu et~al.(2018{\natexlab{a}})Wu, Wan, Yue, and
  Keutzer]{wu2018squeezeseg}
Bichen Wu, Alvin Wan, Xiangyu Yue, and Kurt Keutzer.
\newblock Squeezeseg: Convolutional neural nets with recurrent crf for
  real-time road-object segmentation from 3d lidar point cloud.
\newblock In \emph{2018 IEEE International Conference on Robotics and
  Automation (ICRA)}, pages 1887--1893. IEEE, 2018{\natexlab{a}}.

\bibitem[Wu et~al.(2018{\natexlab{b}})Wu, Zhou, Zhao, Yue, and
  Keutzer]{wu2018squeezesegv2}
Bichen Wu, Xuanyu Zhou, Sicheng Zhao, Xiangyu Yue, and Kurt Keutzer.
\newblock Squeezesegv2: Improved model structure and unsupervised domain
  adaptation for road-object segmentation from a lidar point cloud.
\newblock \emph{arXiv preprint arXiv:1809.08495}, 2018{\natexlab{b}}.

\end{thebibliography}

\end{document}